\documentclass[10pt,twocolumn,letterpaper]{article}

\usepackage{cvpr}
\usepackage{booktabs}
\usepackage{times}
\usepackage{epsfig}
\usepackage{graphicx}
\usepackage{amsmath}
\usepackage{amssymb}
\usepackage{subcaption}
\usepackage{dsfont}
\usepackage{float}
\usepackage{footmisc}
\usepackage[font=small,textfont=it]{caption}
\usepackage[toc,page]{appendix}

\usepackage[pagebackref=true,breaklinks=true,letterpaper=true,colorlinks,bookmarks=false]{hyperref}

\cvprfinalcopy 

\def\cvprPaperID{2960} 

\ifcvprfinal\fi

\usepackage{xcolor}
\newcommand{\MS}[1]{\textcolor{red}{MS: #1}} 
\newcommand{\coco}[0]{MS-COCO}
\newcommand{\pascal}[0]{\textsc{Pascal} VOC}
\newcommand{\pascalplus}[0]{\textsc{Pascal}3D+}

\begin{document}

\title{Straight to Shapes: Real-time Detection of Encoded Shapes}

\author{Saumya Jetley
\footnote[1]{}\\
\and
Michael Sapienza
\thanks{Authors contributed equally}
\thanks{M. Sapienza performed this research at the University of Oxford, and is currently with Samsung Research America, Mountain View CA.}
\and
Stuart Golodetz
\and
Philip H.S. Torr\\
\and
Department of Engineering Science\\
University of Oxford\\
{\tt\small {\{sjetley,smg,phst\}}@robots.ox.ac.uk, m.sapienza@samsung.com}
}

\maketitle

\begin{abstract}
\vspace{-2mm}
Current object detection approaches predict bounding boxes that provide little instance-specific information beyond location, scale and aspect ratio.
  In this work, we propose to regress directly to objects' shapes in addition to their bounding boxes and categories.
 It is crucial to find an appropriate shape representation that is compact and decodable, and in which objects can be compared for higher-order concepts such as view similarity, pose variation and occlusion.
  To achieve this, we use a denoising convolutional auto-encoder to learn a low-dimensional shape embedding space. We place the decoder network after a fast end-to-end deep convolutional network that is trained to regress directly to the shape vectors provided by the auto-encoder.
  This yields what to the best of our knowledge is the first real-time
  shape prediction network, running at ~$35$ FPS on a high-end desktop.
  With higher-order shape reasoning well-integrated into the network pipeline, the network shows the useful practical quality of generalising to unseen categories that are similar to the ones in the training set, something that most existing approaches fail to handle.
\end{abstract}

\vspace{-\baselineskip}

\section{Introduction}

Automatically detecting \cite{Girshick2015} and delineating \cite{Dai2015} object instances in images is a core problem in computer vision, with wide-ranging applications.
For example, knowing the individual boundaries of nearby objects can allow robots to grasp them \cite{Rao2010}.
To help visually-impaired people become more independent, specially-designed glasses can highlight the boundaries of objects with which to interact \cite{Hicks2013}.

For these applications, real-time frame processing is crucial, and the information provided by bounding box predictions \cite{Ren2015} or a pixel-wise semantic segmentation of the scene \cite{Zheng2015} is not enough.
A bounding box captures no more than the location, scale and aspect ratio of an object, 
a fairly coarse representation that provides no information about the object's boundary.
On the other hand, bottom-up pixel-labelling approaches have no explicit notion of local and global object shape.
To achieve adherence to local boundaries and impose spatial and appearance consistency in the semantic space, recent works \cite{chen2016deeplab,Zheng2015} post-process using conditional random fields (CRFs).
Such post-processing quickly becomes intractable for the higher-order constructs required to capture
object structure,
pose and occlusions.

\begin{figure}[tb]
  \centering
  \includegraphics[width=0.95\columnwidth]{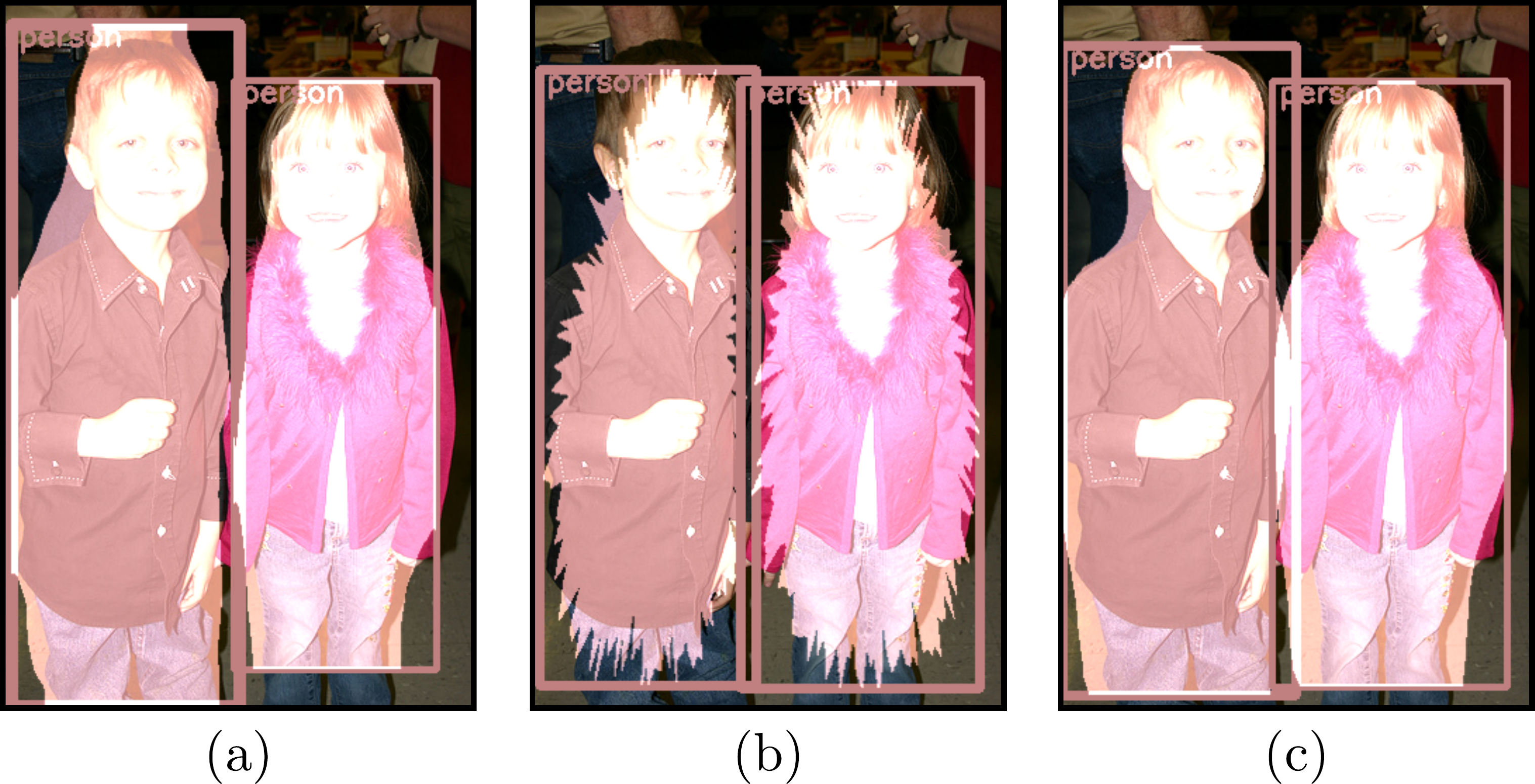}
  \vspace{-2mm}
  \caption{Testing direct regression to shape on a validation image from \pascal\ \cite{Everingham2005} using
           \textbf{(a)} binary shape masks,
           \textbf{(b)} radial vectors, and
           \textbf{(c)} a learned shape embedding.
           }
  \label{fig:differentshapeencodings}
  \vspace{-\baselineskip}
\end{figure}

In response to the above-mentioned shortcomings, we implement an embedding space that incorporates notions of object shape, pose and occlusion patterns.
A deep regression network is then trained to map input image patches to this embedding space to tease apart an object's category and its shape mask.
Fig.~\ref{fig:differentshapeencodings} shows an example of our network regressing to three different shape representations.
Observably, object shape and category are correlated for many classes, so training the learner with a single objective function that includes both terms is mutually-reinforcing.

\begin{figure}
  \centering
  \includegraphics[width=0.98\columnwidth]{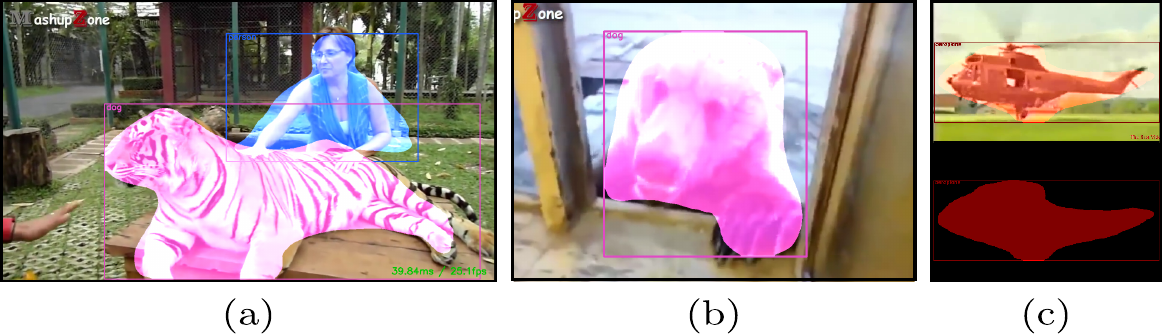}
  \vspace{-2mm}
  \caption{Detecting the shapes of previously-unseen categories in images from YouTube videos:
           \textbf{(a)} a person strokes a tiger,
           \textbf{(b)} a bear is fed through a window, and
           \textbf{(c)} a helicopter crashes into the ground (see supplementary video).
           The tiger and bear are both detected as class `dog', and the helicopter is detected as `aeroplane'.
           Despite wrong categories, the predicted shape masks capture useful information about the types of objects that are present.
          }
  \label{fig:previouslyunseencategories}
  \vspace{-\baselineskip}
\end{figure}

We propose an embedding space that is learnt on the binary instance masks in a class-agnostic manner. 
By design, the embedding space is \emph{compact}, \emph{decodable} (supports mapping of binary instance masks in and out of the space),
\emph{continuous}
(instance masks degrade gracefully around the point of interest in the space)
and \emph{interpretable} (one in which we can reason about the similarities between shapes and their corresponding categories).
Our formulation is thus able to leverage shape reasoning and extend the prediction of shape masks to object categories our network has never seen before,
but which have similar characteristics to the ones seen during training, like the tiger, bear and helicopter in Fig.~\ref{fig:previouslyunseencategories}.

It is also important to note that we tackle the detection of shapes of individual instances
in a top-down paradigm with a single sweep of the network.
This is in contrast to recent works \cite{dai2015instance, Arnab2016} combining top-down and bottom-up paradigms with various sequential arrangements of bounding-box detection,
pixel-wise segmentation and category recognition to achieve instance segmentation.
Due to a sequential processing, the error in the above networks is additive.
For instance, if bounding-box detection comes first and cuts through an
object boundary, pixel-wise labelling inside the box can never retrieve the correct object boundary.
Likewise, if detection follows pixel-wise labelling, it is
difficult to estimate object boundaries for overlapping instances of the same class.

\begin{figure}[!t]
  \centering
  \includegraphics[width=0.98\columnwidth]{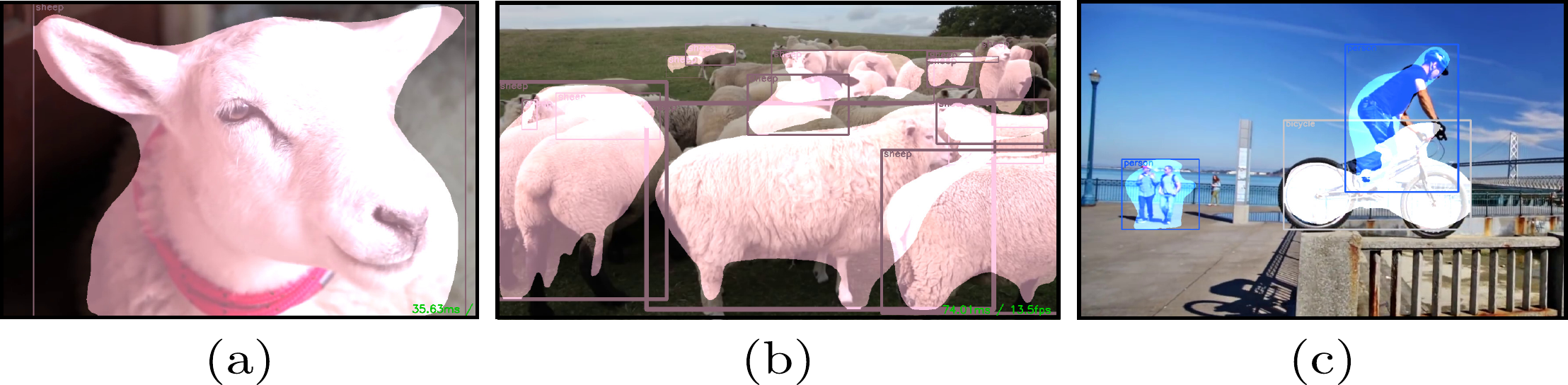}
  \vspace{-2mm}
  \caption{Regressing to single or multiple instances with large intra-category shape variation:
           consider e.g.\ the shape of \textbf{(a)} a sheep's head, in contrast to \textbf{(b)} its body.
           In \textbf{(c)}, we detect multiple overlapping shapes, in this case a person on a bike.
           }
  \label{fig:multipleshapes}
  \vspace{-\baselineskip}
\end{figure}

\textbf{Contributions.}
The proposed approach overcomes the aforementioned difficulties by
directly regressing to multiple object locations, shapes, and categories, as shown in Fig.~\ref{fig:multipleshapes}.
Crucial to the process is the learning of a compact and decodable embedding space in which shapes can be described and compared.
To this end, we demonstrate the use of a denoising auto-encoder~\cite{vincent2008extracting} in encoding real-world shape templates.
Moreover, the single-sweep processing affords our implementation real-time capabilities.
To the best of our knowledge, this is the first real-time shape prediction network, running at $35$ FPS on a high-end desktop with an i7-4960 Processor (3.6GHz, 12-core) and a Titan X GPU.
Our shape prediction network thus builds upon what is currently possible at the intersection of object detection \cite{Redmon2016},
instance segmentation \cite{Hariharan2014} and joint embeddings \cite{Li2015}.

\section{Related Work}
\label{sec:relatedwork}

\textbf{Importance of shape cues.}
In the past, shape cues have been successfully used to guide object recognition \cite{belongie_malik_objreco_using_shapecon} and localisation \cite{jain1996object}, by virtue of their ability to capture discriminative, category-specific details.
With the advent of deep networks, suitable features best adapted to the task are assumed to be learnt by the network during training~\cite{lecun1995convolutional}, 
by optimising an appropriate loss function.
In existing object detection pipelines, this loss is computed on the 4D bounding box specifications \cite{Redmon2016},
ignoring the use of shape to guide learning.
By contrast, regressing to detailed shape vectors affords the network a more informed supervision. 

\textbf{Predicting object shape masks.}
Predicting object shape masks has previously been identified as an important practical challenge \cite{Ferrari2009,Marszalek2007}. However, these approaches contain independently-optimised processing blocks and have shown limited success, working only for images containing only a single object and/or with high background contrast.  More recently, Pinheiro \etal\ \cite{Pinheiro2015,Pinheiro2016} have explored the practical benefits of learning to predict object proposals as pixel-wise segmentation masks.
However, their networks are composed of disjoint stages for predicting segmentation masks, object locations and object classes.
By contrast, and in a spirit similar to YOLO~\cite{Redmon2016}, the architecture that we propose for predicting shape embeddings (from which the shape masks can be reconstructed) is simpler and contains only one network, with a single objective.

\textbf{Extending bounding box detectors to shape.}
Recent progress in bounding box object detection has been fuelled by the representational power of deep networks.
R-CNN \cite{Girshick2014} made massive leaps in detection accuracy by using a deep network to predict the category of pre-generated bounding box proposals.
Soon after, Faster R-CNN~\cite{Ren2015} emerged. It does away with separate stages for region proposal generation and classification, instead using a region proposal network that shares convolutional features with a detection network, achieving benefits in computational efficiency and accuracy.
More recently, the YOLO \cite{Redmon2016} detection network streamlined object detection by replacing the separate proposal generation and classification stages with regression to spatially-separated bounding boxes and class probabilities. This resulted in real-time object detection rates.
Inspired by YOLO, we extend the network to predict shape encodings, achieving similar speeds, but with the advantage of incorporating additional shape information.



\textbf{Shape embeddings.}
The simplest embedding space we explore in this work consists of appropriately down-sampled binary masks.
Other shape representations are also possible starting points for establishing this space;
a comprehensive review of shape representations can be found in \cite{Zhang2004}.
Moreover, inspired by the way in which \cite{Mikolov2013} learned word embeddings to find a feature space in which similar words are close together,
we aim to learn a \emph{shape} embedding in which to reason about the shape of objects.
In contrast to \cite{Mikolov2013}, however, we seek a lower-dimensional encoding (compared to the dimensions of a binary shape mask) to make detection by regression feasible.
We observe the compression, noise-handling and reconstruction capabilities of the learned shape encodings to be significantly superior to the hand-crafted shape representations.

\textbf{Learned shape embeddings.}
Due to the recent successes of learned latent representations for small binary images of digits and faces \cite{Hinton2006},
we learn a shape embedding by training an auto-encoder.
The Shape Boltzmann Machine (SBM) of~\cite{eslami2014shape} uses a deep restricted Boltzmann architecture to model a shape space that is realistic and generalisable.
However, the above networks are trained to model shapes and are not optimised for reconstruction.
Furthermore, it is important to note that the fully-connected construction of SBM restricts it to small synthetic images with a resolution of around $50 \times 50$.
To circumvent this practical limitation, we incorporate convolutional layers, and train the model as a denoising auto-encoder~\cite{vincent2008extracting}.
Thus, in a first, we extend the full benefit of auto-encoders to real-world shape templates.
Interestingly, by learning a continuous and decodable shape embedding,
we are able to predict masks for categories that have not been observed previously but share close shape similarity with those in the train set (see Fig.~\ref{fig:previouslyunseencategories}).
We are able to achieve this performance at a fraction of the dimensionality of hand-crafted shape representations.

Our work is also related to that of Li \etal\ \cite{Li2015},
which handles single image-to-embedding mappings \cite{Tekin2016} and assumes that objects in the input image are prominent. By contrast, we regress to multiple shape embeddings, thereby extending the approach to object detection (see Fig.~\ref{fig:multipleshapes}).
Another important distinction is that the shape embeddings in \cite{Li2015} are hand-crafted and non-decodable, preventing the synthesis of new shapes from arbitrary coordinates in the embedding space and limiting the applications to retrieval with nearest-neighbour search.


\textbf{Potential applications: instance segmentation and occlusion handling.}
Recent works \cite{Gupta2014,Dai2015} on instance segmentation attempt to combine the benefits of the top-down and bottom-up paradigms by performing pixel-wise class labelling within each bounding box prediction.
Such bounding box-based approaches err primarily for occluded objects and their boundaries.
A bottom-up approach like~\cite{chen2015multi} resorted to post-hoc object-level or category-specific reasoning to combine the separated parts of object instances.
Also noteworthy is the work by Dai \etal\ \cite{Dai2016}, which extends a fully-convolutional network (FCN) \cite{Long2015} to predict position-sensitive instance score maps that then need to be assembled into object instances at the output.
Another approach \cite{RomeraParedes2015} makes use of a recurrent neural network (RNN) architecture,
which is trained with a single loss function in order to predict object instances sequentially.
Although this has shown promise on a limited set of object categories,
it has difficulty remembering all past predictions, and therefore finding those that it needs to predict next.


Ultimately, describing an object as a set of pixels with an associated category has its limitations.
In particular, shapes represented as pixel sets are hard to compare: there is no obvious way of computing \emph{meaningful} distances between high-dimensional object masks.
This is unfortunate, because an ability to compare shapes is extremely useful: the shape of an object is not merely an abstract property, but can indicate something fundamental about the object's capabilities and role.
For example, whilst there is significant variety amongst the many legged animals in the world, they are clearly more similar to each other than to an inanimate object such as a lorry, and the shapes involved reflect this.
As such, if we can find a
representation in which we are able to compare shapes quantitatively, we can use the distance values to provide important clues about objects,
even for categories on which we have not trained.
For these reasons, 
we tackle the shape prediction problem as part of an end-to-end pipeline, using an intermediate embedding space that knows about realistic category-level poses, occlusions and shape priors, as learnt from the training data.


\section{Proposed Approach}

\begin{figure*}
  \centering
  \includegraphics[width=0.95\textwidth]{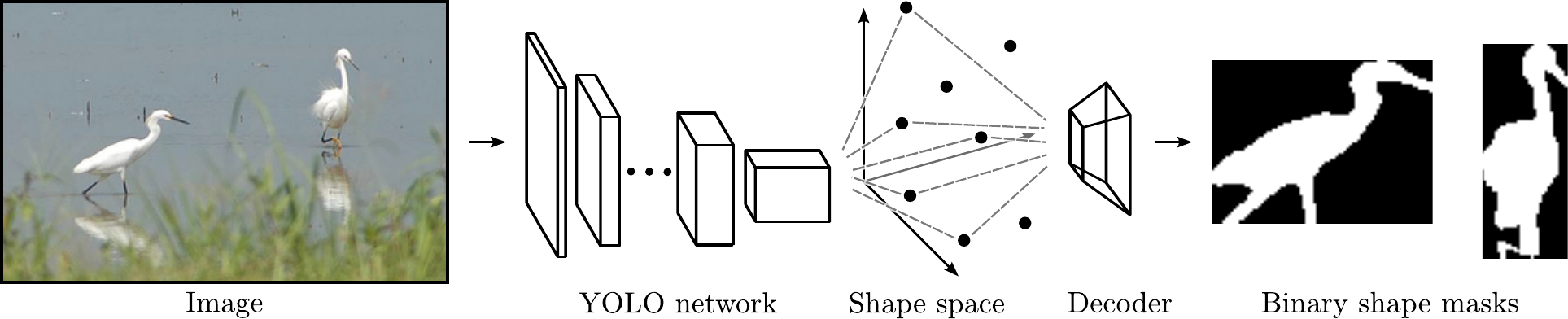}
  \vspace{-2mm}
  \caption{From left, an input image is rescaled and passed through a single network that regresses directly to multiple object hypotheses (category, location, aspect and shape).
  The learned shape encoding is compact, decodable, continuous and interpretable,
  allowing the prediction of shapes not seen in the training data, and more meaningful distance comparisons between shapes.
  Note that predicting encoded shapes enables higher-order reasoning about the pose of the bird (upright, facing right) and how it compares to other birds: this would be impossible from a bounding box alone.
  Moreover, the current shape space is learned in an unsupervised manner and any annotations superimposed on the shape space would add to the predicted information at test time.
  }
  \label{fig:pipeline}
  \vspace{-\baselineskip}
\end{figure*}

\textbf{Deep regression network.}
An object's shape is one of its fundamental properties, which, when coupled with its location, scale and category,
gives rich information about its possible pose and coarse depth, and how one can interact with it.
To take advantage of this, we therefore extend the state-of-the-art YOLO \cite{Redmon2016} object detection network that we mentioned in \S\ref{sec:relatedwork} to regress to not only object locations,
confidence scores and conditional probabilities for each category, but also detailed shape encodings, as illustrated in Fig.~\ref{fig:pipeline}.
The details of how we do this are described in \S\ref{sec:deepregressionnetwork}.

\textbf{Decodable shape representation.}
We require a shape embedding that is both compact and decodable for the object shape prediction application.
Moreover, we would benefit immensely from a representation that embodies higher-order understanding of object shape, realistic poses and occlusion patterns.
For instance, we would like the distances between shapes in the representation to reflect our own understanding of the similarity between shapes.
To this end, in \S\ref{sec:decodableshaperepresentation} we investigate three decodable shape representations:
i) downsampled binary masks,
ii) radial vectors and
iii) learned shape encodings.

In \S\ref{sec:experiments}, we join the ideas from \S\ref{sec:deepregressionnetwork} and \S\ref{sec:decodableshaperepresentation},
and evaluate the merits of our real-time shape detection pipeline.

\section{Deep Regression Network}
\label{sec:deepregressionnetwork}

This section draws inspiration from the YOLO detection algorithm \cite{Redmon2016} and extends it for the prediction of shapes.
Like YOLO, our network reasons globally about objects in the image and is trained to minimise a single objective function.
However, we augment YOLO's object representation with an encoded vector that denotes an object's shape.

The YOLO pipeline starts by dividing the input image into an $S \times S$ grid.
If the centre of an object lies within a cell, then that cell becomes responsible for detecting that object.
Here, each grid cell predicts $B=2$ sets of boxes, confidence scores and shapes with a dimensionality of $N$ per set, as well as a probability mass function.

The box prediction corresponds to the $4$-coordinates that fully specify the tightest fitting bounding box to the object.
The confidence score is calculated as the intersection over union (IoU) between the predicted and target box, if a target box exists, and zero otherwise.
Each shape encoding represents the shape of an object, independent of its location and aspect ratio.
For instance, in order to calculate the binary shape mask that we provide as a target for learning, we take the ground truth object segmentation mask, binarise it and rescale it as part of the encoding process.
Overall, if the shape is to be encoded by a $16 \times 16$ binary mask, then $N = 1 + 4 + 256$.

Each grid cell also predicts a conditional probability mass function, which, when multiplied by the object confidence score,
results in a score reflecting the confidence for the category and its overlap:
\begin{equation}
  p(c|o) * p(o) * IoU = p(c) * IoU.
\end{equation}
As input to our network, we feed in $448\times448$ images,
which are transformed using random rotations, translations, spatial scaling and pixel scaling.
The network's target is augmented by transforming the ground truth shapes with the same geometric transformations as above.
The final size of the target vector for a single image becomes
\begin{equation}
  D = S \times S \times (N \times B + |C|),
\end{equation}
where $|C|$ is the number of dataset categories, $N$ includes parameters for the confidence score, minimum bounding box and shape encoding, and $S \times S$ is the total number of cells in the image grid.

For training and inference, we use Darknet \cite{Darknet2013}.
We optimise the sum of squared errors between the predicted network output and a target tensor.
The four-component loss function used to train the network can be expressed as
$  L = L_{\textnormal{box}}
            + L_{\textnormal{conf}}
            + L_{\textnormal{shape}}
            + L_{\textnormal{pmf}},
$
where 
\begin{equation}
  L_{\textnormal{shape}} =
              \lambda_{\textnormal{shape}}
              \sum_{i=0}^{S^2}
              \sum_{j=0}^{B}
              \sum_{k=0}^{N-5}
              \mathds{1}^{obj}_{ij}
              \left( \tau_{ijk} - \hat{\tau}_{ijk} \right)^2.
\end{equation}
Here, $\tau$ is the target shape representation and $\hat{\tau}$ is the predicted shape representation.
Definitions for the other three loss components can be found in \cite{Redmon2016}.

Since a prediction only requires one forward pass through the network,
it is very fast at test time (\S\ref{sec:experiments}).
We use non-maximal suppression to reduce the number of predicted overlapping shapes.

\section{Decodable Shape Representation}
\label{sec:decodableshaperepresentation}

The shape embedding space is a crucial part of the proposed formulation. We believe that the tasks of constructing the embedding space and learning a mapping from the image domain to this embedding space are intrinsically independent and should therefore be treated separately. Moreover, doing so makes the two tasks more tractable \cite{Li2015}. We thus experiment with two hand-crafted representations and one learned shape representation to establish the embedding space, described as follows. 

\subsection{Downsampled binary shape masks}
A very simple shape descriptor can be obtained just by downsampling the full-size binary shape mask.
Given a fixed descriptor size $d = k \times k$, we downsize the image using OpenCV's \texttt{INTER\_AREA} resampling approach.
For reconstruction, the descriptor can be resized back up again using bicubic interpolation.

\subsection{Radial representation}
A radial descriptor represents a shape as a series of distances between some centre point within the shape and points deterministically distributed over the shape's boundary.
We choose the boundary points by finding where rays cast outwards from the centre point at angles uniformly distributed over $[0,2\pi)$ intersect the boundary.
To improve shape reconstruction it makes sense to choose a centre point that has a direct line of sight to as much of the boundary as possible.
In practice,
we construct several radial descriptors for each shape and pick the one with maximal IoU.
More details can be found in the supplementary material.

\subsection{Learned shape encoding}

The aim is to implement a network that can compress an input binary mask to a comparatively lower-dimensional space.
Autoencoders have already been shown to improve over PCA and logistic PCA at a similar task for digits, curves and faces~\cite{Hinton2006}.
Thus, in an initial experiment, we use the fully-conneced autoencoder architecture (784-1000-500-250-30) of \cite{Hinton2006} to learn a shape embedding space for the Caltech-101 silhouettes contained in the trainval set.
In addition to the restricted Boltzmann machine-style pre-training of~\cite{Hinton2006}, we fine-tune the network weights using cross-entropy error between the predicted and target shape masks. This setup yields a reconstruction IoU of $0.85$ on the test set. It becomes evident that this architecture, also employed in~\cite{eslami2014shape,NIPS2012_4774}, is not able to exploit the spatial redundancy in visual data due to its fully-connected nature; for the same reason, it is not scalable to large images.  

\begin{figure}[tb]
  \centering
  \includegraphics[width=0.6\columnwidth]{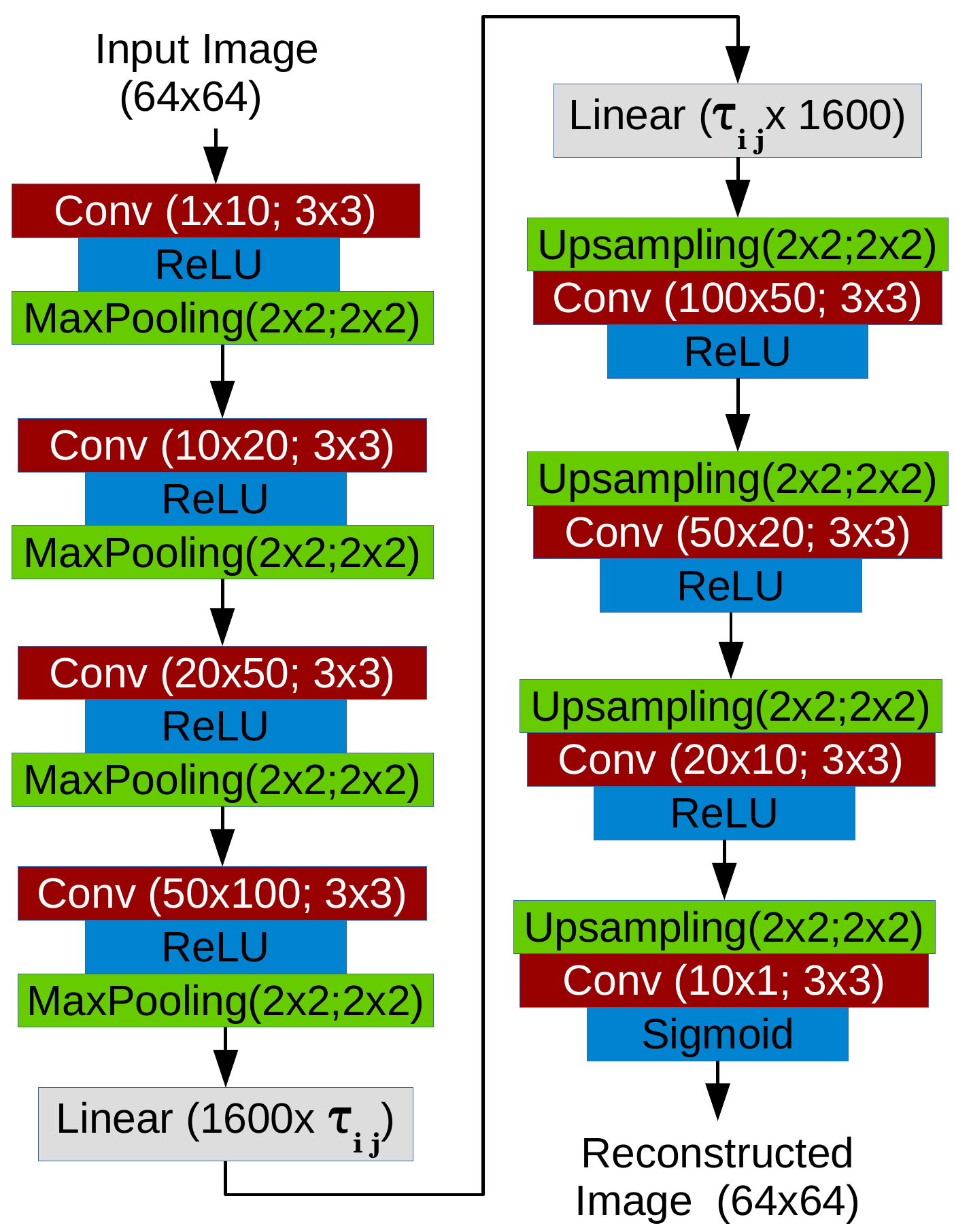}
  \vspace{-2mm}
  \caption{Block diagram of the denoising autoencoder using convolutional layers. The learnt shape representations are used to train the regression network.}
  \label{fig:denoisingAE}
  \vspace{-\baselineskip}
\end{figure}

It is important to note that similar shape patterns or boundary formations can be found at different locations in the input image: for example, the front and rear tyre of a bicyle or motorbike look similar. We can leverage this repetition of visual patterns by the use of a convolutional construct. Accordingly, a block diagram of our autoencoder network is shown in Fig.~\ref{fig:denoisingAE}. To prevent the network from learning an identity mapping to a higher-dimensional space in the initial layers of the network meant for data-disentangling, we introduce white noise in the input image so the network generalises to a denoiser~\cite{vincent2008extracting}. We minimize the binary cross-entropy loss to adjust the network weights. It can be simplified as
\begin{equation}
L_{bce}=\sum_{i=1}^{P} \begin{cases}
-log(\hat{p}_i) & \text{if  \hspace{2pt}} p_i = 1 \\
-log(1-\hat{p}_i) & \text{if \hspace{2pt}} p_i = 0,
\end{cases}
\end{equation}
where $P$ is the total number of pixels in the image, $\hat{p}_i$ is the predicted probability and $p_i$ is the pixel value.

\section{Experiments and Results}
\label{sec:experiments}


Our method allows us to unify detection and segmentation by integrating higher-order concepts of shape into the end-to-end prediction pipeline. 
We thus study the application of our proposed prediction network to instance segmentation (\S\ref{subsec:instancesegmentation}),
where we found that having the concept of shapes allows it to predict shape masks even for unseen categories.
Section \ref{subsec:zeroshotsegmentation} presents a comparative analysis of this zero-shot segmentation capability.
Essential to our formulation is the need to reconstruct accurate shape masks from an interpretable and low-dimensional shape space.
We thus also analyse the trade-offs between three different shape spaces in \S\ref{subsec:shapereconstructionerror}.
 
\subsection{Comparing the Shape Representations}
\label{subsec:shapereconstructionerror}
The choice of representation is crucial, as it dictates what aspects of a shape are made explicit and the ease with which it can be manipulated \cite{Marr1982}.
We thus compare the shape representations outlined in \S\ref{sec:decodableshaperepresentation} on three different axes: i) their ability to accurately represent shapes at low dimensionalities, ii) the continuity of the associated shape space and iii)~the structure of the shape space.

\begin{figure}[tb]
  \centering
  \includegraphics[width=0.75\columnwidth]{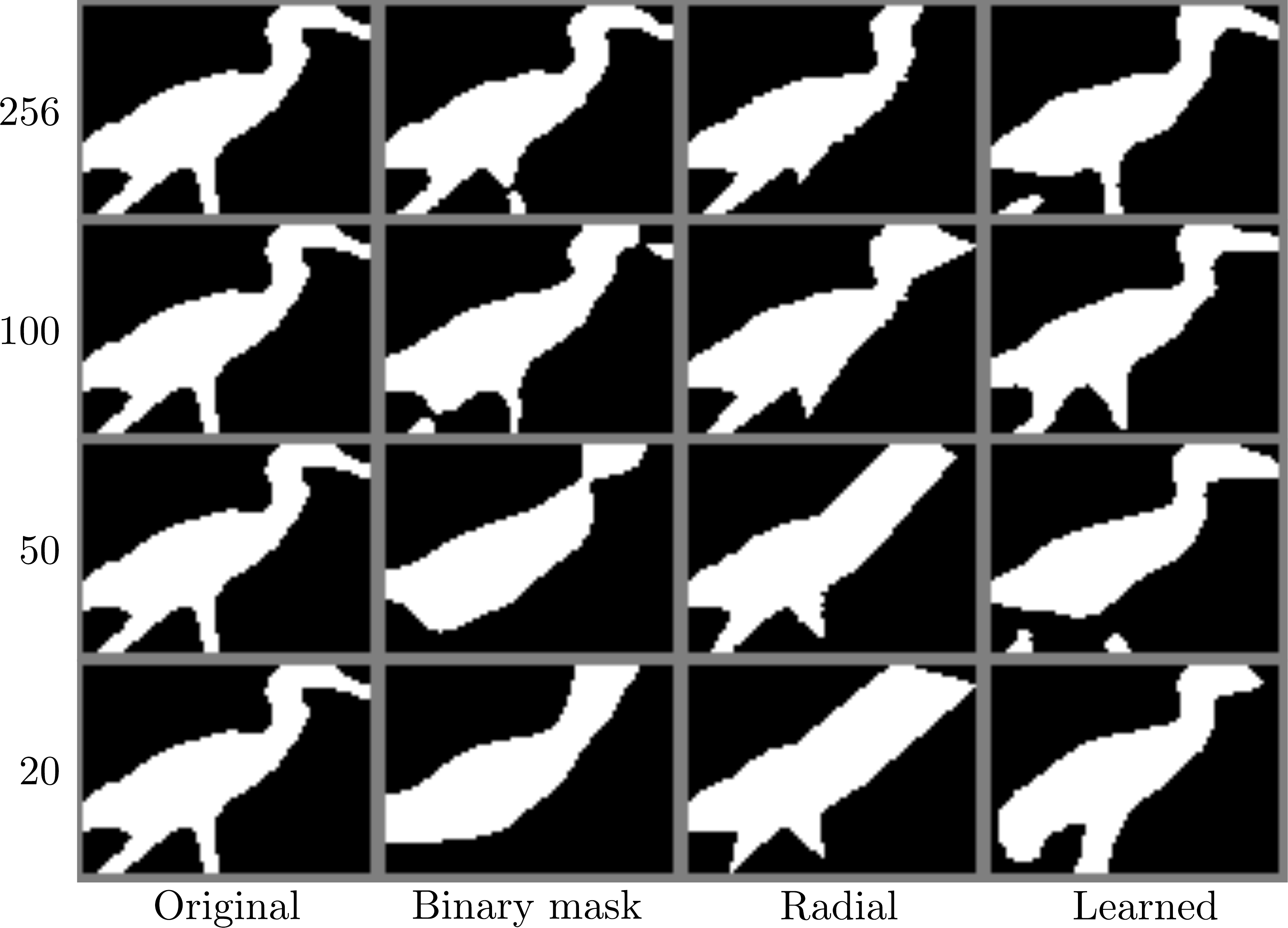}
  \vspace{-2mm}
  \caption{
  As the dimensionality reduces, both binary masks and radial vectors lose precious details, such as the legs and head of the bird, whilst the learned embedding is able to preserve the overall topology of the shape and degrades more gracefully.}
\label{fig:reconstructionerrorvsdim}
\end{figure}

\noindent \textbf{Representation ability:} We compute the average shape reconstruction error (one minus the IoU) yielded by each of our three descriptors at different sizes on the SBD dataset. 
%
%
%
Table~\ref{tab:reconstructionerror} shows how the error for each descriptor varies with the representation size.
At low descriptor dimensionalities our encodings are better able to preserve shape than either of the two hand-crafted alternatives.
However, the benefits fade away at larger dimensionalities.
A qualitative example of the artifacts introduced by lowering the dimensionality of each shape representation can be seen in Fig.~\ref{fig:reconstructionerrorvsdim}: both of the two hand-crafted representations experience a severe decrease in quality as the size decreases, whereas our representation still manages to capture the major topological features of the shape even at size $20$.
\begin{table}[!t]
  \centering
  {\footnotesize
  \scalebox{0.95}
  {
  \begin{tabular}{cccccc}
    \toprule
    \textbf{SBD} & 20 (25) & 50 (49) & 100 & 200 (196) & 256 \\
    \midrule
    Downsampled mask & 0.15 & 0.10 & \textbf{0.07} & \textbf{0.05} & \textbf{0.04} \\
    Radial & 0.13 & \textbf{0.08} & \textbf{0.07} & 0.06 & 0.06 \\
    Shape encoding & \textbf{0.125} & \textbf{0.08} & \textbf{0.07} & 0.06 & 0.06 \\
    \bottomrule
  \end{tabular}
  }
  }
  \vspace{1mm}
  \caption{Average shape reconstruction error (one minus the IoU) achieved by each of our three descriptors at different sizes on the SBD validation set~\cite{Everingham2005,Hariharan2011}.
The unbracketed numbers indicate the sizes used for the radial and shape encoding descriptors; the bracketed numbers indicate the sizes used for the downsampled mask descriptors, which are square by design.}
  \label{tab:reconstructionerror}
  \vspace{-0.5cm}
\end{table}

\begin{figure}[tb]
  \centering
  \includegraphics[width=0.75\columnwidth]{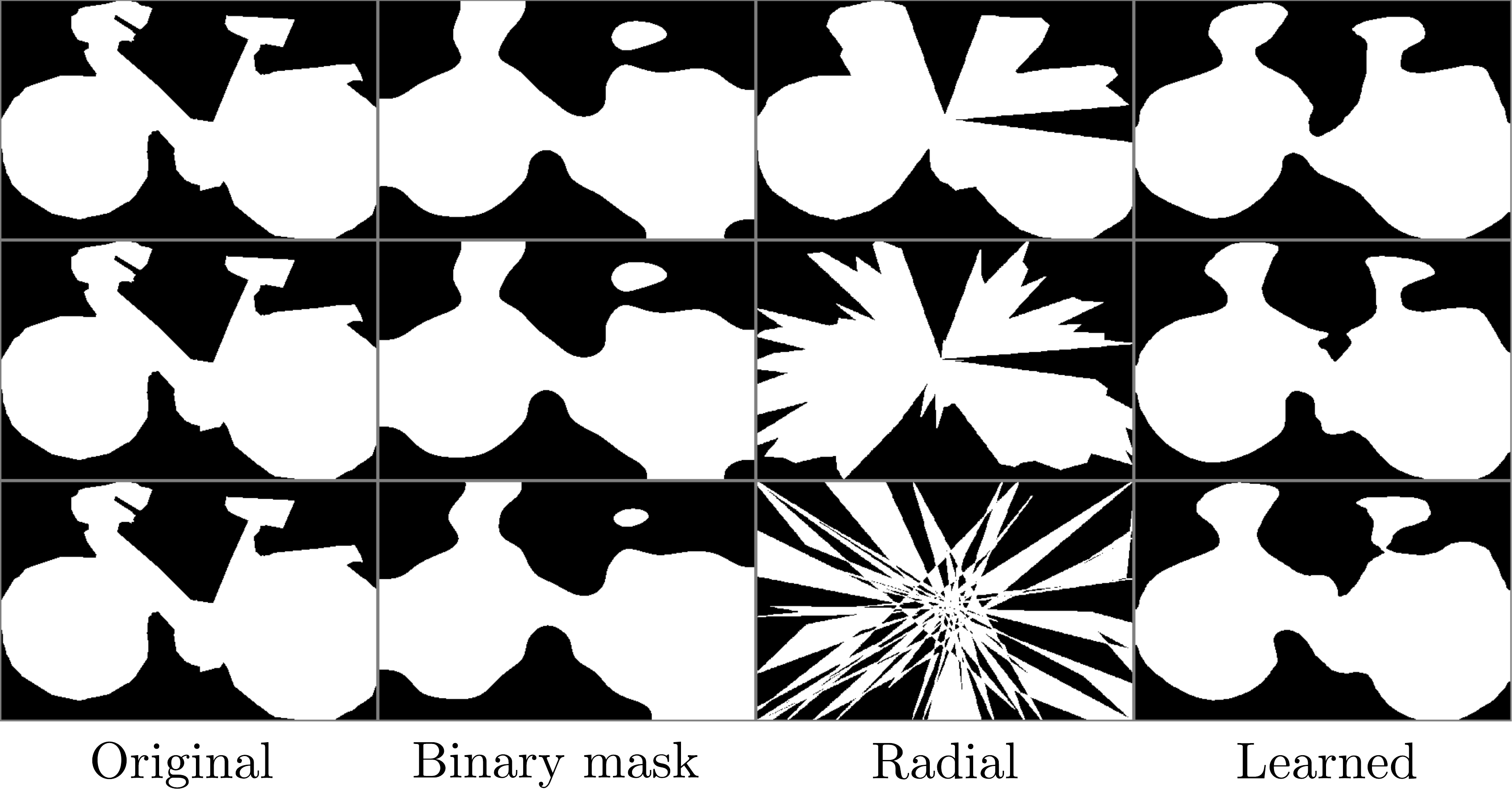}
  \vspace{-2mm}
   \caption{The effect of adding Gaussian noise of zero mean and increasing variance to each of the shape representations.}
  \label{fig:reconstuctionerrorvsnoise}
  \vspace{-\baselineskip}
\end{figure}

\noindent \textbf{Shape space continuity:} To study the continuity of the shape spaces, we add Gaussian noise to the encodings and visualise its effects on the reconstructed shape masks in Fig.~\ref{fig:reconstuctionerrorvsnoise}.  As expected, an increase in noise degrades all representations; however, a little bit of noise in the radial vector distorts the shape beyond recognition. The binary mask loses precision at the boundaries, as the wheels of the bicycle blend into the border of the image. The learned shape representation is able to preserve details such as the seat and the circular wheels better than the hand-crafted alternatives.

\begin{figure*}[tb]
  \centering
  \includegraphics[width=0.95\textwidth]{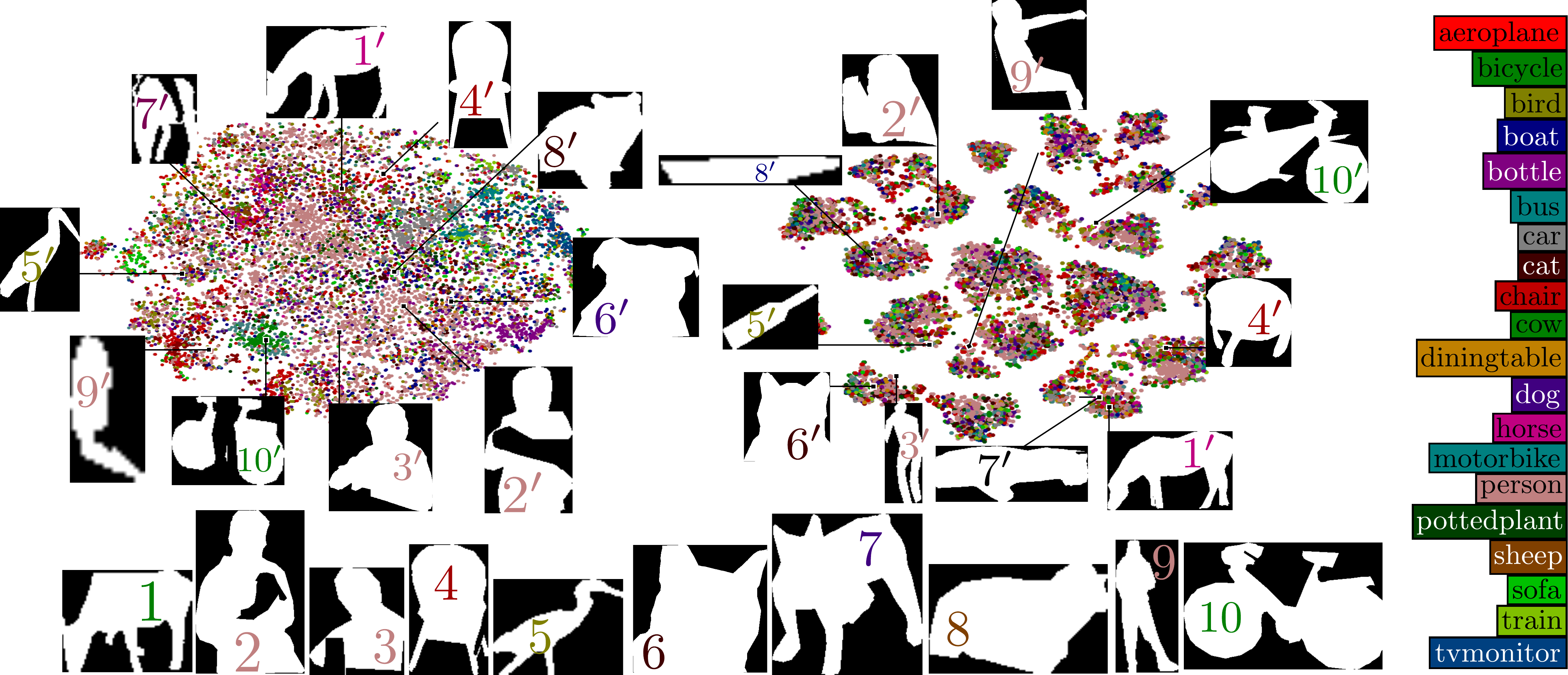}
  \vspace{-2mm}
  \caption{Visualising the 20D learned embedding space (left) next to the 256D radial vector space (right) with t-SNE \cite{VanDerMaaten2008}. 
Notice the differences between the anchors' nearest neighbours (NNs) in the two representations. For instance, consider the bicycle anchor (no.\ 10), which is facing left: its NN in the learned embedding is also facing left (even though it is occluded), whilst its NN in the radial space is facing to the right.
         }
  \label{fig:visualisingembedding}
  \vspace{-\baselineskip}
\end{figure*}

\noindent \textbf{Shape space structure:} We visualise the structure of the shape spaces associated with two representations -- the learnt encodings and the radial descriptors -- in Fig.~\ref{fig:visualisingembedding}. Even at a first glance, it is obvious that the shape space for the learnt encoding provides a better category-based clustering of the shape masks. For instance, as we move from the top left region of the space to the bottom right along a diagonal, we observe a transition from samples with large boundary projections (horses and birds) to upright shapes of humans and finally to more streamlined shapes such as bottles. There is no such clear structure in the shape space for radial descriptions: indeed, the categories seem to be scattered across various disparate clusters.
We also visualise the nearest neighbour of each of a pre-selected set of $10$ `anchor' images, where the 
learnt embedding space observably shows a closer match between neighbouring shapes.

Following this qualitative analysis, we use the \pascalplus\ \cite{Xiang2014} dataset to quantify the difference in shape space structure by gathering object pose and category statistics around neighbourhoods in the respective shape spaces.
We divide the \pascalplus\ dataset into a train set of 11578 images and a val set of 120 images.
The val set is formed by randomly selecting 10 images from each of the 12 rigid object categories of \pascalplus\ that are a subset of the 20 \pascal\ categories.
For each val set image, we find the 50 nearest neighbours in the train set and calculate the variance in pose and category with respect to the validation image.
The averaged results over all val images and categories are shown in Table~\ref{tab:shapespacestructure}.

We find that poses (elevation, azimuth, distance) of objects in the vicinity of a shape in the learned shape encoding space are more similar to each other than those in the radial space (smaller mean average variance - mAV).
Moreover, the majority category amongst neighbouring objects is more likely (80\% of the time) to match to the category of the central shape.
These numbers demonstrate that the shape space is more interpretable than the radial space.
\begin{table}[!t]
  \centering
  {\footnotesize
  \scalebox{0.95}
  {
  \begin{tabular}{ccc}
    \toprule
    & \textbf{Radial (20D)} & \textbf{Shape Encoding (20D)} \\
    \midrule
    mAV (azimuth)   & 1.163 & 1.008 \\
    mAV (elevation) & 0.294 & 0.246 \\
    mAV (distance)  & 21.537 & 17.968 \\
    \# GT class = majority class & 7 \text{out of} 12 & 12 \text{out of} 12\\
    mA\% majority class & 0.547 & 0.800 \\
    \bottomrule
  \end{tabular}
  }
  }
  \vspace{1mm}
  \caption{
    Comparing the interpretability of the radial representation with that of the learned shape encoding on the \pascalplus\ dataset.
    The statistics are obtained from the 50 nearest neighbours of a validation shape in each respective space, and then averaged over all 10 validation samples and 12 object categories.
  }
  \label{tab:shapespacestructure}
  \vspace{-0.2cm}
\end{table}

\subsection{Instance Segmentation as an Application}
\label{subsec:instancesegmentation}
The hypothesis that we would like to test is that we are able to regress directly to an object-based representation of the scene.
For this experiment, we train a YOLO-style network that maps directly from a single image to a collection of object masks and their locations
that can be compared against those generated by state-of-the-art instance segmentation methods \cite{Hariharan2014,Arnab2016,Dai2016}.

\textbf{Dataset, splits and performance metrics:}
\label{subsec:datasets}
We train and evaluate our method on the established \pascal\ with SBD annotations \cite{Everingham2005,Hariharan2011} dataset, 
with the same split sets as used by \cite{Hariharan2014,Dai2016}.
We report the mean average precision at various overlap thresholds (mAP$^r$, mAP$^r_{vol}$) \cite{Hariharan2014}.

\textbf{Results and discussion:}
In Table~\ref{tab:instancesegmentation}, we report the instance segmentation results we obtain when regressing to the following shape representations:
i) downsampled binary shape masks ($d=256$),
ii) radial vectors ($d=256$),
and iii) learned embeddings ($d \in \{20,50\}$).
The dimensionalities were chosen based on the reconstruction errors obtained in Table~\ref{tab:reconstructionerror},
where we observed that we are able to use lower-dimensional shape codes with the learned embedding than with the hand-crafted ones (see Fig.~\ref{fig:reconstructionerrorvsdim}).
This reduction in the number of parameters per shape means that we can train neural network models with fewer parameters, and with less chance of overfitting.
This intuition is consistent with our empirical findings in Table~\ref{tab:instancesegmentation}, where we achieve better performance as the dimensionality of the embedding is reduced from $50$ to $20$.
In comparison to both of our embeddings, both downsampled $16 \times 16$ binary masks and $256D$ radial descriptors perform worse, despite having more space available in which to represent the shape.
One possible reason for this in the case of radial descriptors is their higher sensitivity to noise, as can be seen in Fig.~\ref{fig:reconstuctionerrorvsnoise}.
Qualitative results showing the performance of our method on the SBD dataset can be seen in Fig.~\ref{fig:qualitativeresults},
and in the supplementary material.

It will be seen in Table~\ref{tab:instancesegmentation} that our instance segmentation results lag somewhat behind current state-of-the-art methods; this parallels the lag in object detection results experienced by YOLO with respect to state-of-the-art methods in that field.
We hypothesise that this is because both approaches struggle to precisely localise certain objects, especially small ones (see Figure~\ref{fig:qualitativeresults}).
However, like YOLO, our approach runs in real time, something that cannot be achieved by any of the \emph{offline} methods against which we compare (as can be seen in Table~\ref{tab:instancesegmentation}, existing methods are in no way real time, taking at least an order of magnitude longer to run).
We are also much better than the state-of-the-art instance segmentation method of Arnab \etal\ \cite{Arnab2016} at recognising unseen categories (see Fig.~\ref{fig:comparisontostateoftheart}), as discussed in the next section.

\begin{figure}[!t]
\centering
\includegraphics[width=0.8\columnwidth]{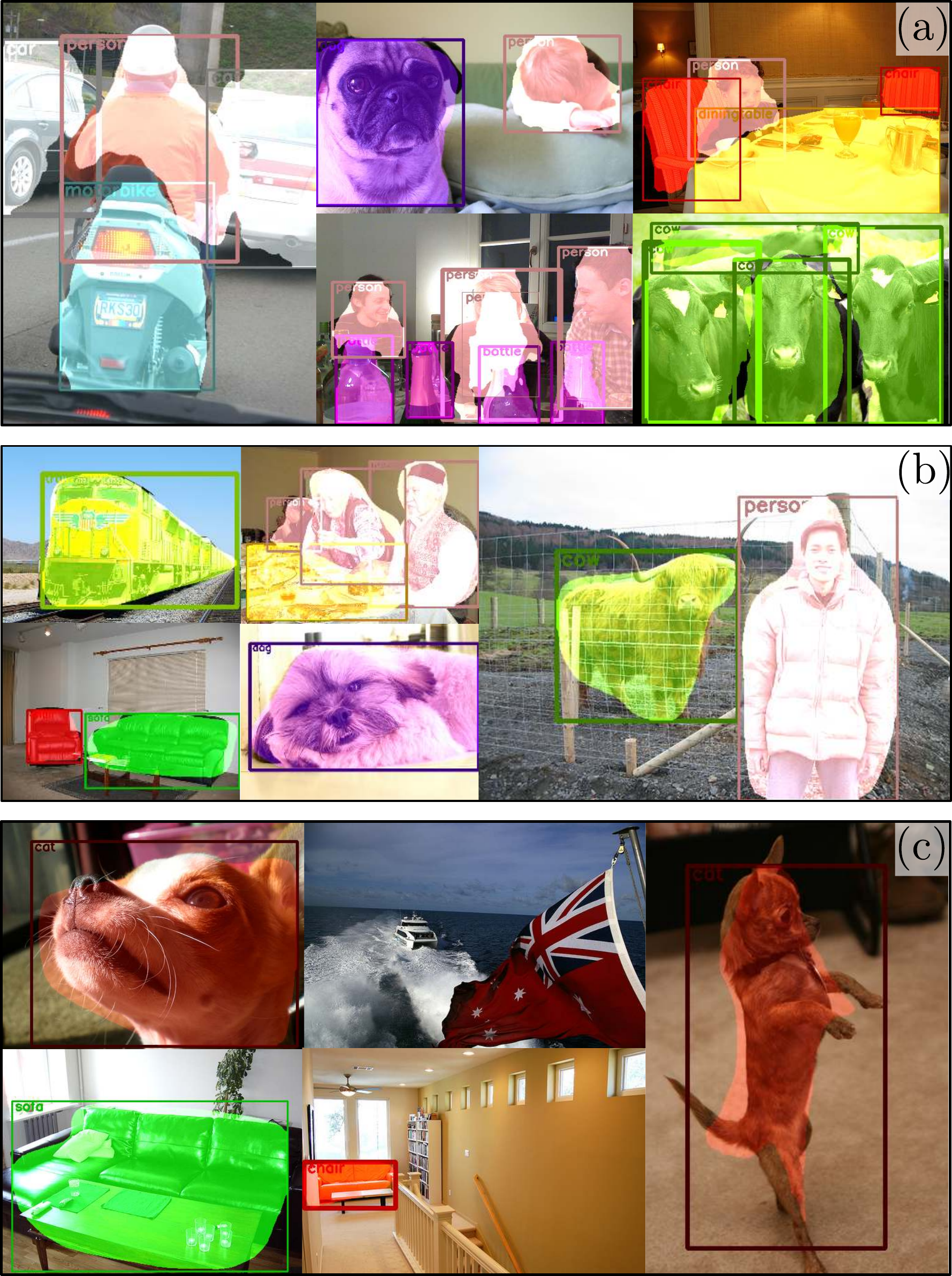}
\vspace{-2mm}
\caption{\textbf{(a)} Correct predictions using downsampled binary masks.
         \textbf{(b)} Correct predictions using 20D learnt shape encodings.
       In the $3^{\text{rd}}$ image, the horns of the cow are missed and the human shape mask gets elongated due to an incorrect bounding box prediction.
     \textbf{(c)} Missed detections using the 20D shape encodings. The network misses out or false fires on small objects ($2^{\text{nd}}$ column). The dogs in the images are falsely categorised as cats, and the sofa incorrectly includes the nearby dining table.}
\label{fig:qualitativeresults}
\vspace{-0.8cm}
\end{figure}

\begin{table}[!t]
  \centering
  {\footnotesize
  \scalebox{0.85}
  {
  \begin{tabular}{lccccc}
    \toprule
                   &\multicolumn{3}{c}{\textbf{SBD (5732 val images)}}   &  & \textbf{Time} \\
                   &\scriptsize{mAP$^r$@.5} & \scriptsize{mAP$^r$@.7} & \scriptsize{mAP$^r_{vol}$}&  & ms \\ \midrule
    BinaryMask     & 32.3             & 12.0             & 28.6          &     & \textbf{26.3} \\
    Radial         & 30.0             &  6.5             & 29.0          &     & 27.1 \\
    Embedding (50) & 32.6             & 14.8             & 28.9          &     & 30.5 \\
    Embedding (20) & \textbf{34.6}    & \textbf{15.0}    & \textbf{31.5} &     & 28.0 \\ \midrule
    SDS \cite{Hariharan2014}        
    & 49.7             & --                & 41.4          &   & 48k \\
    MNC \cite{Dai2016}
    & 65.0             & 46.7             & -- &  & 330 \\
    \bottomrule
  \end{tabular}
  }
  }
  \caption{Quantitative instance segmentation results on PASCAL SBD 2012 val. 
           The timing results were obtained on a high-end desktop containing a Titan X.
         }
  \label{tab:instancesegmentation}
  \vspace{-\baselineskip}
\end{table}

\subsection{Zero-shot segmentation}
\label{subsec:zeroshotsegmentation}
\begin{figure}[tb]
  \centering
  \includegraphics[width=0.95\columnwidth]{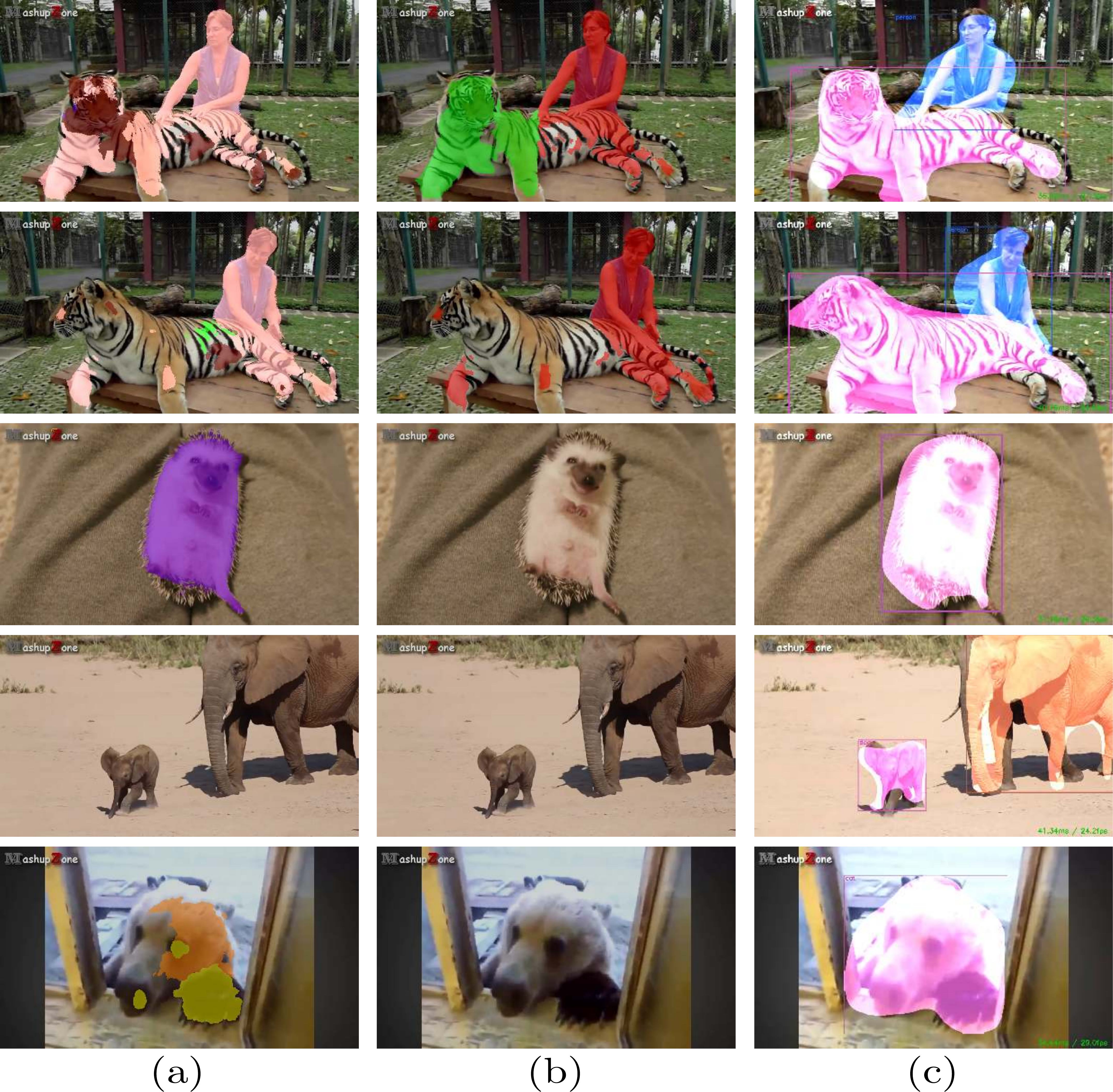}
  \vspace{-2mm}
  \caption{A comparison between the state-of-the-art \textbf{(a)} semantic segmentation~\cite{arnab2016higher}, \textbf{(b)} instance segmentation~\cite{Arnab2016}, and \textbf{(c)} our shape detection results, on images from YouTube videos of animals that are not present in the \pascal\ training set.
    In the first two rows, instance segmentation predicts that the legs of the tiger are human. Our method is more consistent over the tiger images taken from the same video.
    In the lower rows, the instance segmentation approach \textbf{(b)} fails to predict any segments, whilst our method predicts class `dog' for the tiger, hedgehog, baby elephant and bear,
    and class `horse' for the large elephant.    
       }
  \label{fig:comparisontostateoftheart}
  \vspace{-\baselineskip}
\end{figure}

Traditionally, the task of zero-shot learning has been achieved through the use of attributes~\cite{romera2015embarrassingly,zhang2016zero}.
A recent work on zero-shot boundary segmentation~\cite{LiECCV2014} follows a similar approach.
By contrast, our method doesn't need an explicit definition of attributes to extend shape segmentation to unseen categories (see Fig.~\ref{fig:comparisontostateoftheart}); shape attributes are implicitly captured in the definition of the shape space.
We are thus able to scale to object categories that have similar shape appearances at test time.
This is very similar to how humans perform.
For instance, someone who has never seen a tiger before would still be able to segment it properly,
even though he/she might compare it to a \emph{cat- or dog-like animal}, and this is precisely how our network approaches the task.

In order to evaluate our segmentation of unseen class objects,
we take our shape prediction model (50D learned encoding) trained on \pascal\ train and val,
and test it on the 60 object categories in \coco~\cite{Lin2014} not present in \pascal.
We filter images from the entire \coco\ dataset such that no image contains an object of a seen category; yielding 8037 images in our test-bed with a total of 24,396 instances of unseen object classes.
We gather all object predictions with a detection score greater than $t=0.05$, since the unseen category scores are likely to be low,
and then run them through the evaluation code provided by~\cite{Lin2014}, which we modified to calculate IoU over shapes instead of boxes.
The results, in Table~\ref{tab:zeroshotsegmentation}, are specially reported for large objects as the \pascal\ dataset used for training predominantly contains objects ($>80\%$ objects) in the large size range as defined in the \coco\ dataset.
Given that the existing methods perform very poorly for segmentation of unseen classes (Fig.~\ref{fig:comparisontostateoftheart}), we report our results as a strong baseline. We hope that these results will help catalyse research in this direction.

\begin{table}[!t]
  \centering
  {\footnotesize
  \scalebox{0.9}
  {
  \begin{tabular}{lccc}
    \toprule
                   &\multicolumn{3}{c}{\textbf{COCO (8037 val images)}}  \\
                   &\scriptsize{mAP$^r$@.5} (all) & \scriptsize{mAP$^r$@.5} (large) & \scriptsize{AR$^r$@.5} (large) \\ 
		\midrule
    Embedding (50) & 3.6             & 7.1             & 23.2          \\
    \bottomrule
  \end{tabular}
  }
  }
  \vspace{-0.3cm}
  \caption{Quantitative zero-shot segmentation results obtained by running our embedding (50) prediction model on the 60 categories of \coco\ not present in \pascal. 
         }
  \label{tab:zeroshotsegmentation}
  \vspace{-0.3cm}
\end{table}



\section{Conclusion}

In this work, we show for the first time that it is possible to regress directly and simultaneously to multiple object representations that incorporate the notion of shape.
We combine ideas from object detection, instance segmentation and low-dimensional embedding spaces to create a real-time system to detect encoded shapes.
One key factor that allows us to do this is the introduction of a shape embedding space that is compact, decodable, continuous and interpretable.
We find that imbuing object detectors with the knowledge of shape allows us to predict plausible masks for previously-unseen categories, allowing us to detect the presence of objects in images for which current state-of-the-art instance segmentation methods perform poorly or fail.

Our next step is to investigate how our shape prediction approach can be extended to cope with a larger variety of shapes and object categories like those in \coco\ \cite{Lin2014}.

\paragraph{Acknowledgements.}
This work was supported by the EPSRC, ERC grant ERC-2012-AdG 321162-HELIOS, EPSRC grant Seebibyte EP/M013774/1 and EPSRC/MURI grant EP/N019474/1.


\begin{appendices}
\newenvironment{stusubfig}[1]
{
        \begin{figure}[#1]
        \begin{center}
}
{
        \end{center}
        \end{figure}
}

\newenvironment{stusubfig*}[1]
{
        \begin{figure*}[#1]
        \begin{center}
}
{
        \end{center}
        \end{figure*}
}

\section{Training the Regressor}

The regression network architecture was based on the YOLO design \cite{Redmon2016},
which has 24 convolutional layers, followed by 2 fully-connected layers;
the convolutional layers were pre-trained on the ImageNet dataset \cite{Russakovsky2015}.
The relative weights between the loss components were set to\linebreak[4]
$\lambda_{shape} = 0.1$,
$\lambda_{box} = 5$,
$\lambda_{obj} = 1$,
$\lambda_{noobj} = 0.5$ and
$\lambda_{class} = 1$.

We trained the network for $500$ epochs on the training set of the \pascal\ SBD splits \cite{Hariharan2014}\footnote{\url{https://github.com/bharath272/sds_eccv2014/blob/master/train.txt}}, which took about 3 days on one Titan X.
We used a batch size of $64$, a momentum of $0.9$ and a weight decay of $0.0005$.
The learning rates for the various batches were set as follows:
\begin{table}[H]
\centering
\begin{tabular}{cc}
\toprule
\textbf{Batch Numbers} & \textbf{Learning Rate} \\
\midrule
0 -- 200 & $0.001$ \\
201 -- 400 & $0.0025$ \\
401 -- 20,000 & $0.005$ \\
20,001 -- 30,000 & $0.0025$ \\
30,001 -- 40,000 & $0.00125$ \\
40,001+ & $6.25 \times 10^{-4}$ \\
\bottomrule
\end{tabular}
\end{table}
To mitigate the effects of a relatively small dataset ($\sim 5k$ training images), we used data augmentation with parameters generated uniformly within the following ranges:
rotation ($\pm 4^\circ$),
translation ($\pm$ 20\% of image size),
scaling ($\pm$ 3\% of image size),
random flipping,
intensity scaling ($\alpha I + \beta$, with $\alpha$ in the range $\pm2$ and $\beta$ in the range $\pm10$).

In the main paper, we compared our results qualitatively to the work of Arnab \etal\ \cite{Arnab2016}, who kindly tested the algorithm in \cite{Arnab2016} on the images that we selected from YouTube videos.
Note that we did not compare to \cite{Arnab2016} quantitatively, as different training and test data was used in \cite{Arnab2016} (including data from the standard \pascal\ dataset \cite{Everingham2005}, the SBD dataset \cite{Hariharan2014}, and \coco{} \cite{Lin2014}).

\begin{figure}[!t]
\centering
\includegraphics[width=\linewidth]{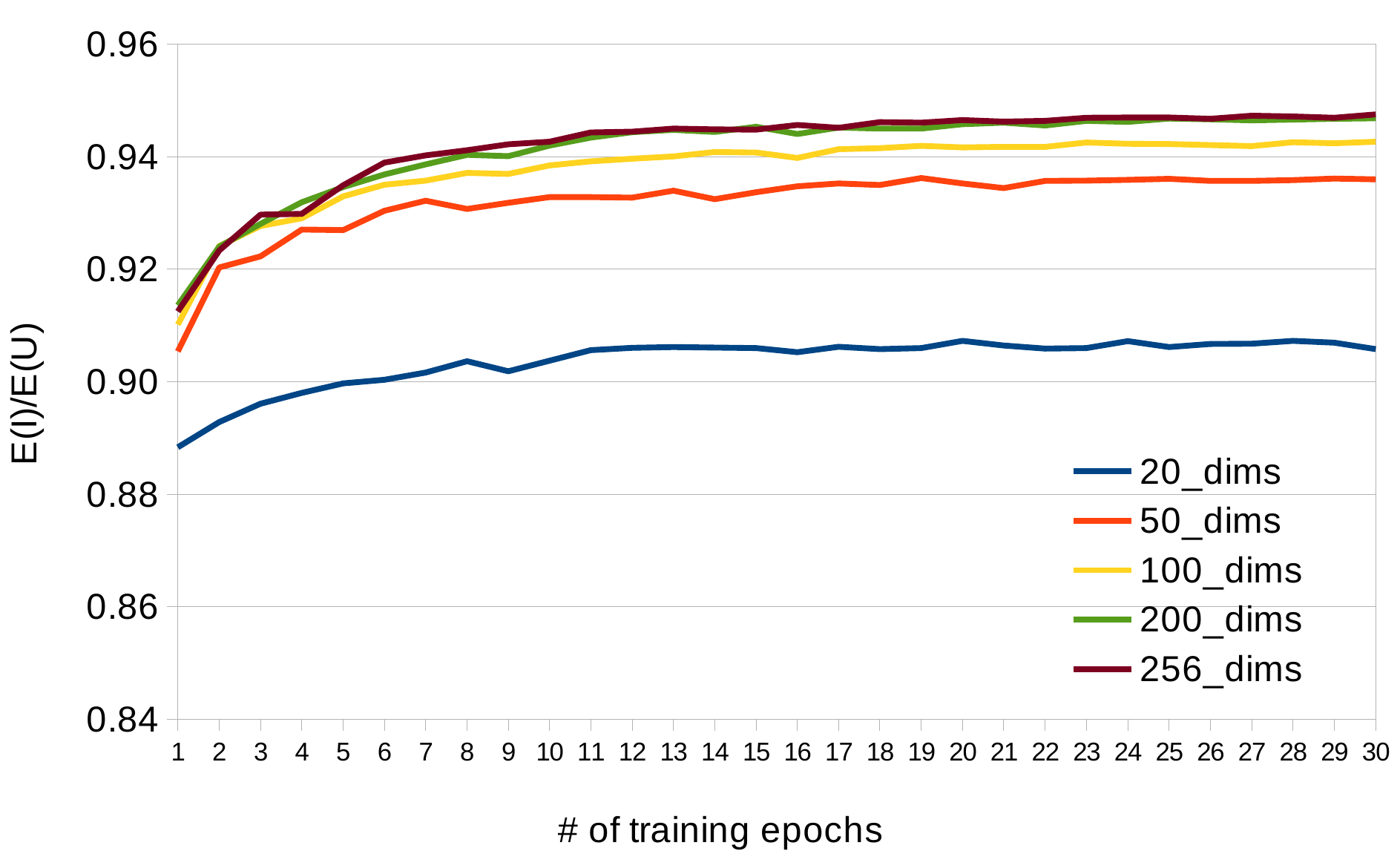}
\caption{Learning curves for the denoising auto-encoder architectures with different sizes of the embedding space.}
\label{fig:conv_plots}
\vspace{-1.5\baselineskip}
\end{figure}

\section{Training the Autoencoder}

We trained the auto-encoder architecture described in \S$5.3$ of the main paper for various dimensionalities of the embedding space, namely $20$, $50$, $100$, $200$ and $256$. For both training and testing purposes, we cropped each image to the boundary of its inset binary mask and resized it to $64 \times 64$. We used a batch size of $128$ and an initial learning rate of $0.001$, and trained the networks for a total of $300$ epochs, with the learning rate being reduced by a factor of $2$ every $60$ epochs. The network weights were tuned using back-propagation and stochastic gradient descent (in particular, the Adam optimization method~\cite{kingma2014adam}).

We adjusted the network weights to minimise the binary cross-entropy loss $\ell_{ce}$ between the input binary mask $m$ and the reconstructed binary mask $m^{recon}$ as follows:
\begin{equation}
\ell_{ce} = -\sum_{i=1}^n m_i \cdot \log(m^{recon}_i) + (1-m_i) \cdot \log(1-m^{recon}_i)
\label{eq:cross_entropy_loss}
\end{equation}
In this, $n$ is the total number of pixels in the input mask.

The convergence plots for various sizes of the embedding space are shown in Fig.~\ref{fig:conv_plots}. It is interesting to note that with the reduction in the size of the embedding space, the performance degrades gracefully and minimally. Noticeably, with a $10$ times reduction in the dimensionality, the expected IoU (Expected(I) / Expected(U)) falls by only around $0.04$ points. 

%
%
\section{Radial Descriptor Computation}

\begin{stusubfig}{!t}
	\begin{subfigure}{.48\linewidth}
		\centering
		\includegraphics[width=\linewidth]{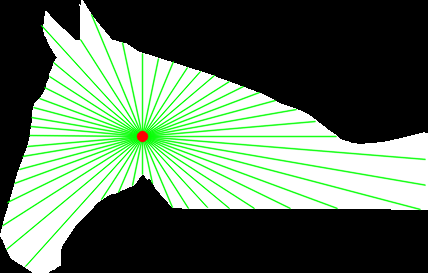}
	\end{subfigure}%
	\hspace{3mm}%
	\begin{subfigure}{.48\linewidth}
		\centering
		\includegraphics[width=\linewidth]{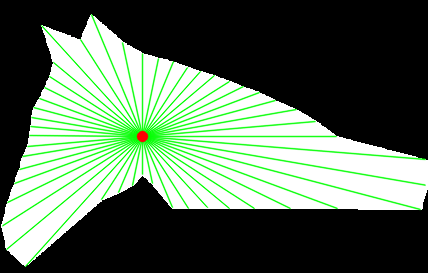}
	\end{subfigure}%
	\\ [3mm]
	\begin{subfigure}{.48\linewidth}
		\centering
		\includegraphics[width=\linewidth]{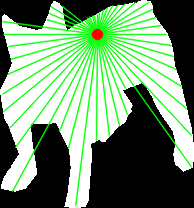}
	\end{subfigure}%
	\hspace{3mm}%
	\begin{subfigure}{.48\linewidth}
		\centering
		\includegraphics[width=\linewidth]{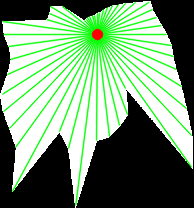}
	\end{subfigure}%
	\\ [3mm]
	\begin{subfigure}{.48\linewidth}
		\centering
		\includegraphics[width=\linewidth]{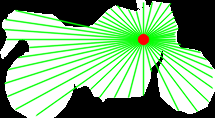}
	\end{subfigure}%
	\hspace{3mm}%
	\begin{subfigure}{.48\linewidth}
		\centering
		\includegraphics[width=\linewidth]{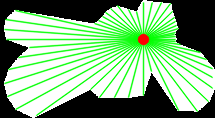}
	\end{subfigure}%
	\\ [3mm]
	\begin{subfigure}{.48\linewidth}
		\centering
		\includegraphics[width=\linewidth]{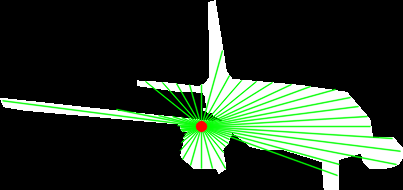}
	\end{subfigure}%
	\hspace{3mm}%
	\begin{subfigure}{.48\linewidth}
		\centering
		\includegraphics[width=\linewidth]{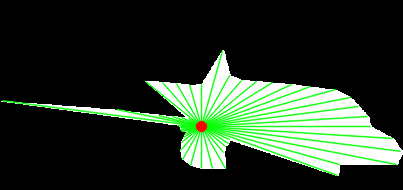}
	\end{subfigure}%
\caption{Some example radial descriptors of size $50$, superimposed on the original masks (left column) and the reconstructed masks (right column). Each chosen centre is shown with a red dot, and we draw green rays between the centre and each point on the represented shape contour. By choosing the centre and the way in which we cast rays appropriately in each case, we achieve much better reconstruction results than we could using the standard centre of the mask and casting rays outwards; however, for more complicated shapes such as the dog (second row) and the aeroplane (fourth row), our radial descriptors fail to capture important details such as the shapes of the ears or the tailplane.}
\label{fig:radialdescriptors}
\vspace{-1.5\baselineskip}
\end{stusubfig}

To compute a radial descriptor for a binary shape mask, we adopt the following scheme.
Given a fixed descriptor size $d$, we allocate the first $2$ elements of the descriptor to contain the $(x,y)$ coordinates of the centre point we choose within the mask, normalised to $[0,1] \times [0,1]$ (where $(0,0)$ denotes the top-left of the mask and $(1,1)$ denotes the bottom-right).
We allocate the remaining $d-2$ elements to contain distances between the chosen centre point and the boundary at evenly-spaced angles in $[0,2\pi)$, each appropriately scaled to fall in the $[0,1]$ range.
In particular, element $i+2$ of the descriptor stores the scaled distance between the centre point and the boundary at an angle of $2\pi i / (d - 2)$.

In practice, to achieve better IoU scores for reconstruction, we construct several radial descriptors for each shape and pick one with maximal IoU.
In particular, we try $25$ possible centre points, evenly spaced on a $5 \times 5$ grid superimposed over the $w \times h$ mask at locations
\[
\left( \frac{jw}{6}, \frac{ih}{6} \right) \; : \; i, j \in [1,5].
\]
We also try conceptually casting rays both away from and towards the centre point (in practice, casting rays towards the centre point is implemented by casting rays away from the centre point and only stopping when the outer boundary of the shape is reached).
This scheme significantly improves the IoU scores we can achieve for reconstruction with respect to always using the centre of the mask and casting rays outwards.

Figure~\ref{fig:radialdescriptors} shows some of the size $50$ radial descriptors we calculate, superimposed on both the original and reconstructed masks so as to illustrate what aspects of shape the descriptors can and cannot capture.
The chosen centre in each case is shown with a red dot, and we draw green rays between the centre and each point on the represented shape contour.

\section{Software Implementation}

\begin{figure}[!t]
  \centering
  \includegraphics[width=0.95\columnwidth]{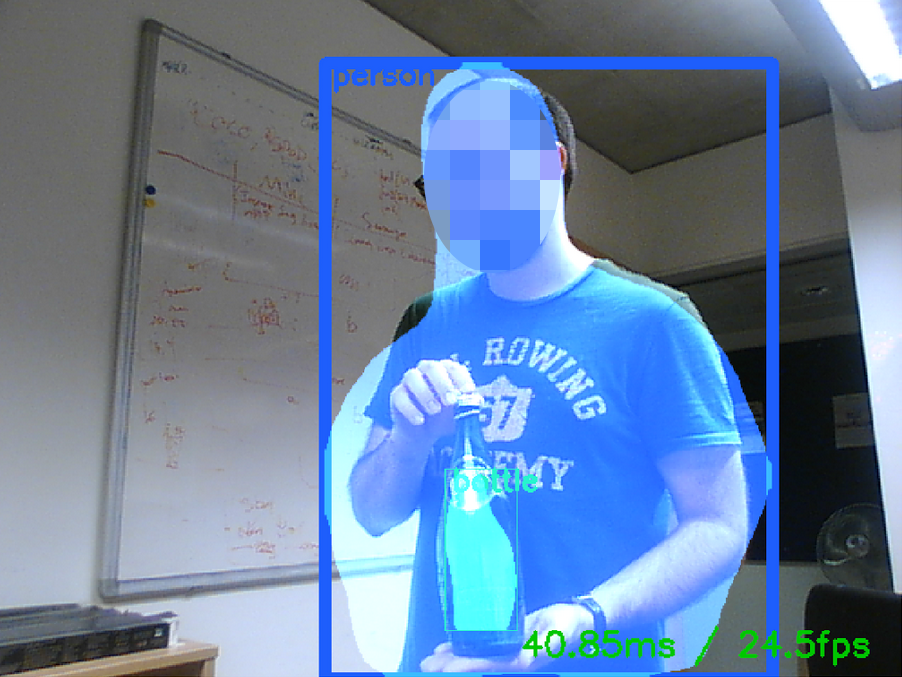}
  \caption{Testing direct regression to shape from a live webcam image stream.}
  \label{fig:webcam}
  \vspace{-\baselineskip}
\end{figure}

We used the Darknet framework \cite{Darknet2013} (written in C) from June 2016 for training our shape prediction model. We developed our own C++ software to deal with dataset loading, manipulation, transformations and preparation for the neural network.
We also developed C++ code to evaluate our system on instance segmentation; our code is several times faster than the standard MATLAB version of Hariharan et al.\ \cite{Hariharan2014}.
For ease of development, we used the Torch framework for learning the shape embedding, and interfaced the C++ and Lua code using the LuaBridge library\footnote{\url{https://github.com/vinniefalco/LuaBridge}}.
This interface includes delays that could be avoided by implementing an optimised version of our Lua code in C++.

\paragraph{Real-Time Demo.}
We also developed some code to demonstrate that our system generalises to scenes captured by a web-camera, as illustrated in Fig.~\ref{fig:webcam}.

\paragraph{Reproducible Results.}
The code and models used to obtain our results are now available at \href{url}{https://github.com/torrvision/straighttoshapes}.

\section{Further Qualitative Results}
\begin{figure*}
\centering
\includegraphics[width=0.7\textwidth]{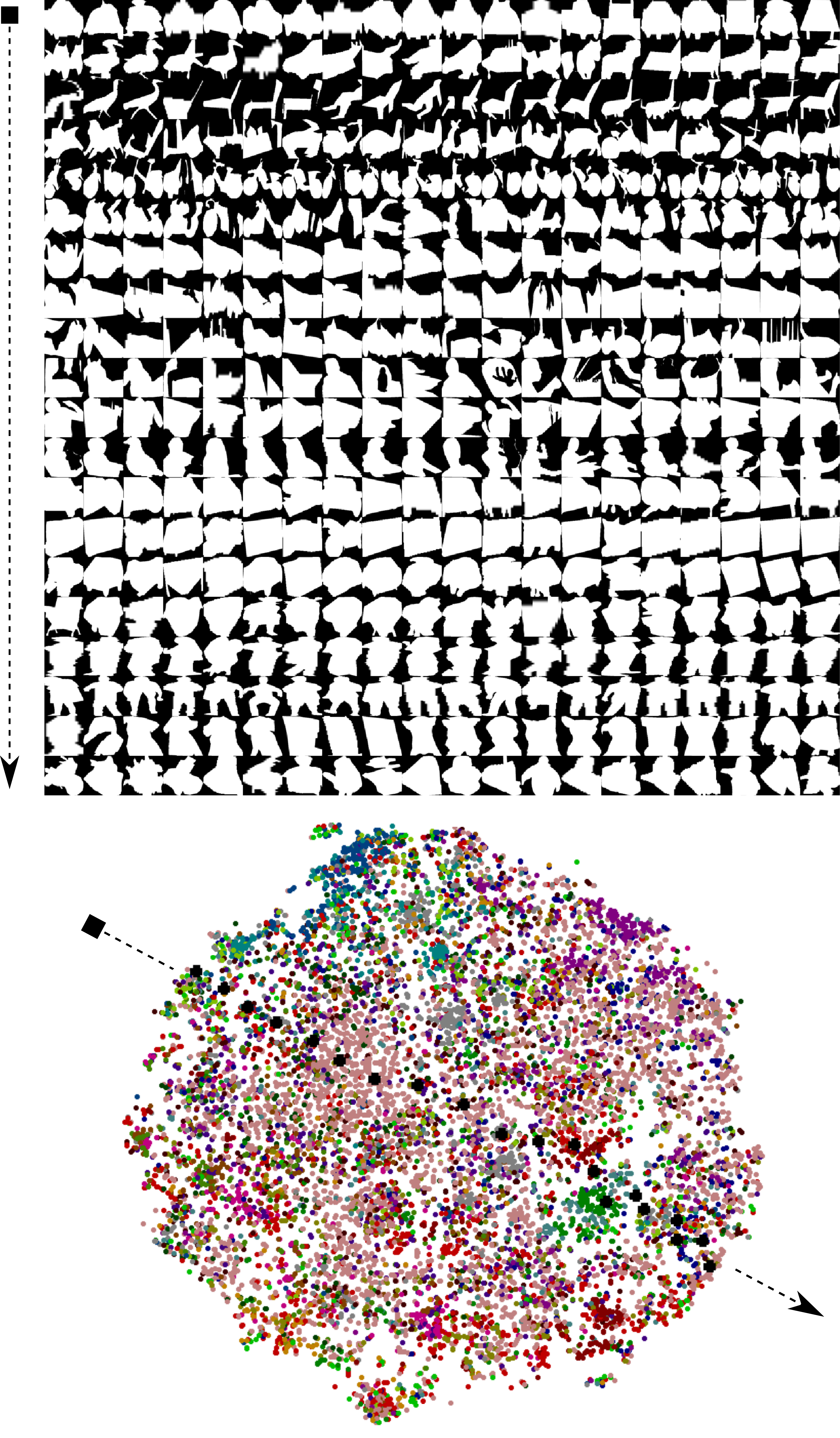}
\caption{
  Visualisation of the $16 \times 16$ binary mask embedding space.
  (top) Ground truth binary masks.
  Rows (in top) represent the $20$ neighbouring binary masks for each of the $20$ anchor points in the embedding space (bottom).
}
\label{fig:binaryMask}
\end{figure*}

\begin{figure*}
\centering
\includegraphics[width=0.7\textwidth]{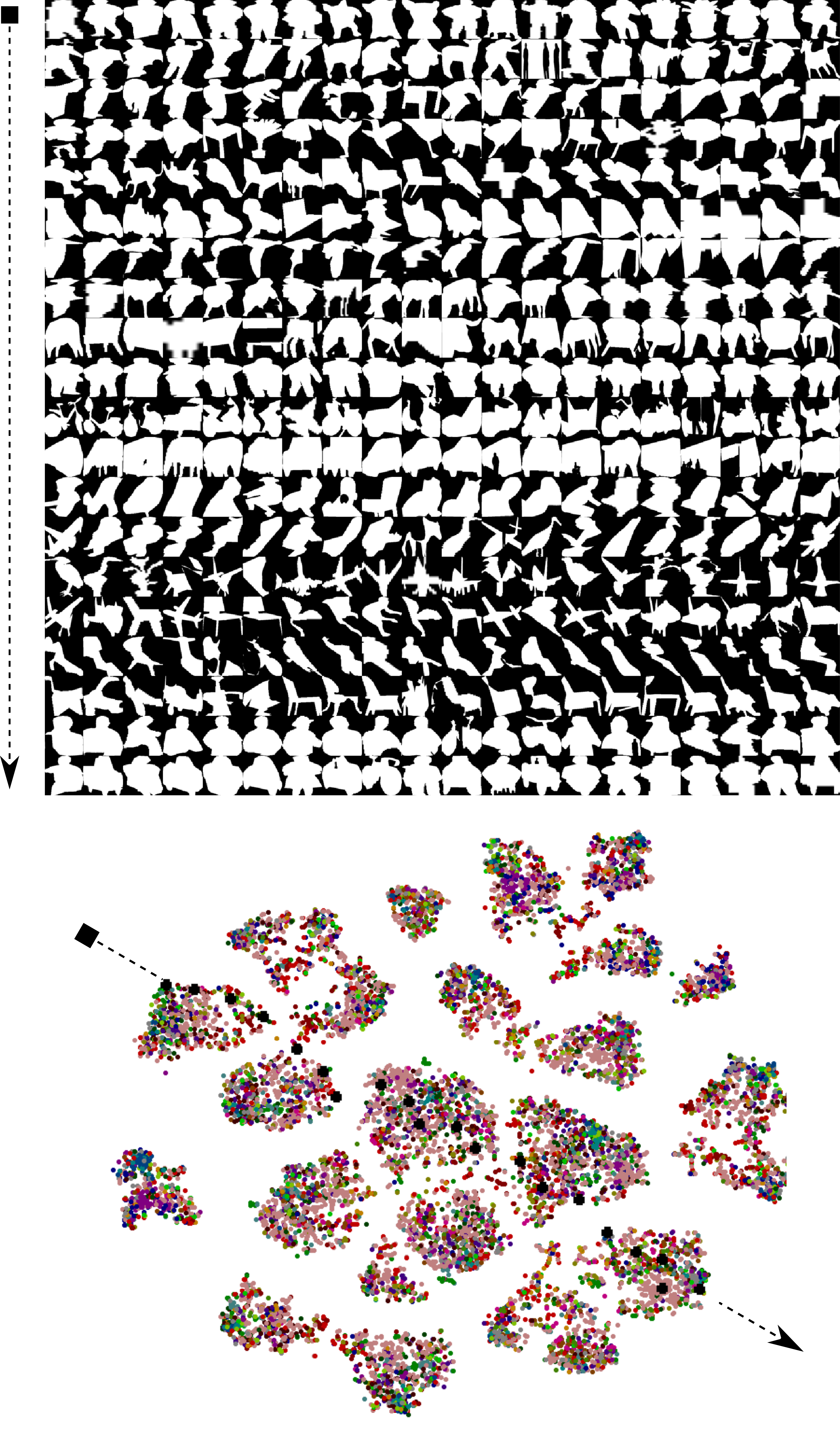}
\caption{
  Visualisation of the $256$-D radial vector embedding space.
  (top) Ground truth binary masks.
  Rows (in top) represent the $20$ neighbouring binary masks for each of the $20$ anchor points in the embedding space (bottom).
}
\label{fig:radialVector}
\end{figure*}

\begin{figure*}
\centering
\includegraphics[width=0.7\textwidth]{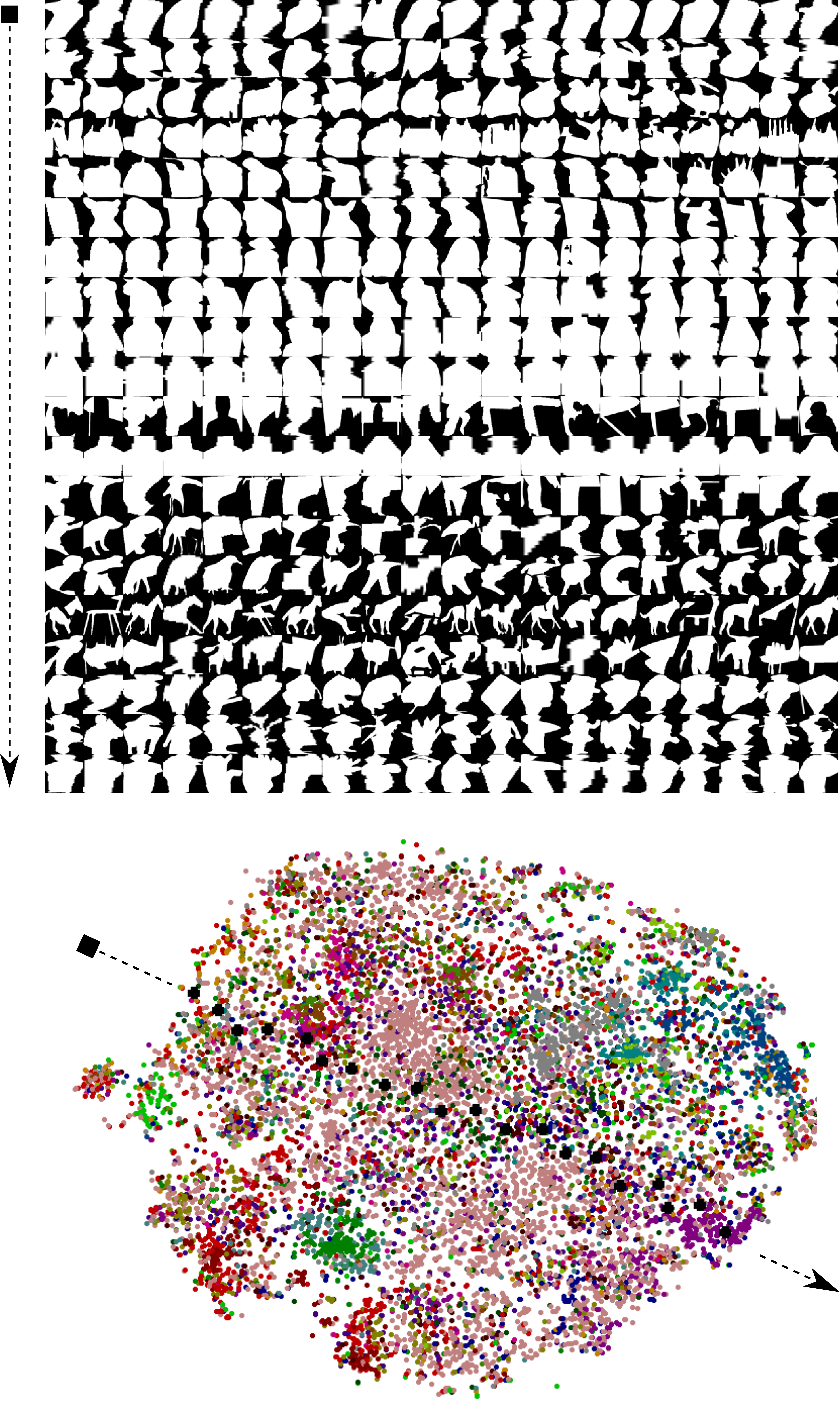}
\caption{
  Visualisation of the $20$-D learned embedding space.
  (top) Ground truth binary masks.
  Rows (in top) represent the $20$ neighbouring binary masks for each of the $20$ anchor points in the embedding space (bottom).
}
\label{fig:learnedEmbedding}
\end{figure*}

\paragraph{Visualisation of the embedding spaces.}
In Figs.~\ref{fig:binaryMask},~\ref{fig:radialVector}, and~\ref{fig:learnedEmbedding}, we visualise the embedding spaces produced by the binary masks, radial vectors and learned embedding respectively.
We fix $20$ anchor points in each of the embedding spaces around the main diagonal of the 2D space.
For each of the anchor points, we sample the $20$ nearest neighbours to observe how the shapes are organised in the space.
At a macroscopic level, as we move along the anchor points (across the rows of the tiled images), we notice that the shapes change between being bulky and rounded to shapes with multiple protrusions.
For the learned embedding space, this pattern is more pronounced (see Fig.~\ref{fig:learnedEmbedding}).
In the radial space, both bulky and thin shapes are present along a single row (see Fig.~\ref{fig:radialVector} rows 3 and 7),
thus in places showing an inferior organisation.

Both the binary mask space and the learned embedding space are well-organised for shape similarity.
It is interesting to note that the learned embedding space achieves a similar shape organisation to the downsampled binary mask space, but with an order of magnitude reduction in size.

\begin{figure*}[!p]
  \centering
  \includegraphics[width=0.95\textwidth]{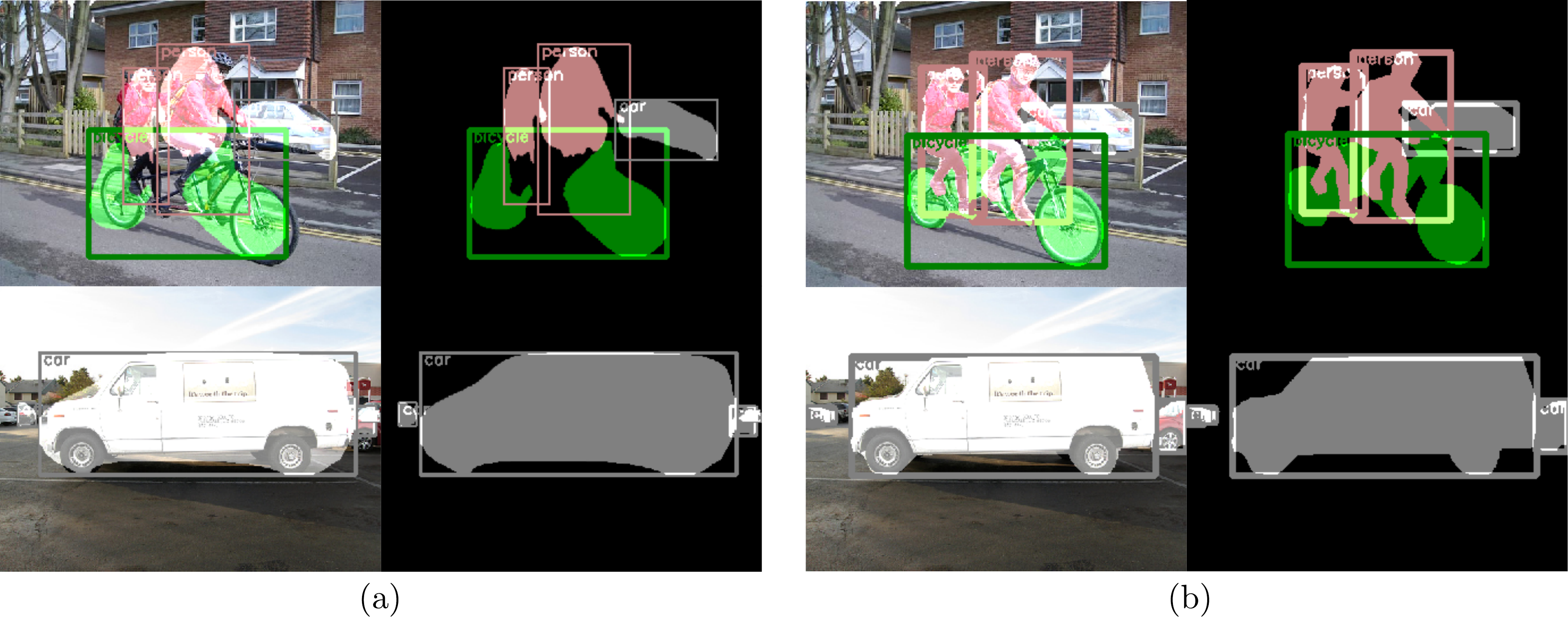}
  \caption{
    A more detailed look at the results. \textbf{(a)} The results from our shape prediction network next to \textbf{(b)} the ground truth.
    \textbf{(a)} (top) Two people ride a tandem bike. Note that our system predicts shapes and bounding boxes (learnt jointly) for localisation.
    As such, any errors in localisation will affect the location at which the shape gets superimposed onto the image (the box does not hug the bike tires tightly).
    In \textbf{(b)} (top), note the complexity of the shapes that need to be predicted.
    In the bottom row, multiple cars are detected. The predicted shape contains information on the direction that the car is facing, information that is not available from bounding boxes alone.
    It is noteworthy that even thought the output shape mask is not pixel-accurate, it still contains a lot of information about the object's orientation and shape, and its similarity to other shapes in the shape embedding.
  }
  \label{fig:moreresults1}
\end{figure*}

\begin{figure*}[!p]
  \centering
  \includegraphics[width=0.95\textwidth]{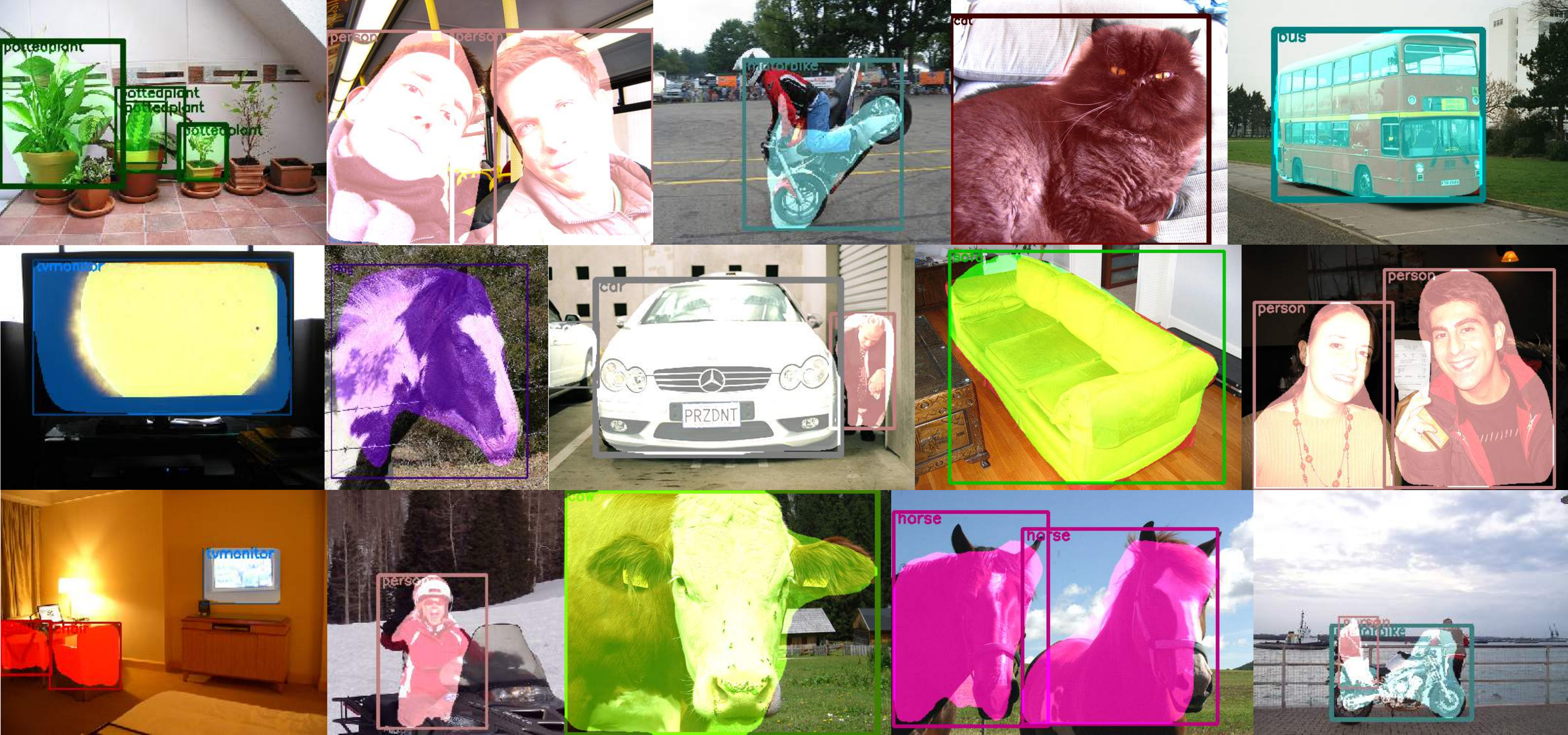}
  \caption{
    Additional qualitative results from the \pascal\ (SBD) validation set.
    Note the huge variety in shapes, even within a single category.
  }
  \label{fig:moreresults2}
\end{figure*}

\paragraph{Shape prediction.}
In Figs.~\ref{fig:moreresults1} and~\ref{fig:moreresults2}, we show additional qualitative results of our shape prediction system on the \pascal\ \cite{Everingham2005} (SBD) validation set.

\end{appendices}

\end{document}